\documentclass[sigconf]{acmart}

\AtBeginDocument{%
  }

\copyrightyear{2024} 
\acmYear{2024} 
\setcopyright{acmlicensed}\acmConference[SIGGRAPH Conference Papers '24]{Special Interest Group on Computer Graphics and Interactive Techniques Conference Conference Papers '24}{July 27-August 1, 2024}{Denver, CO, USA}
\acmBooktitle{Special Interest Group on Computer Graphics and Interactive Techniques Conference Conference Papers '24 (SIGGRAPH Conference Papers '24), July 27-August 1, 2024, Denver, CO, USA}
\acmDOI{10.1145/3641519.3657469}
\acmISBN{979-8-4007-0525-0/24/07}

\usepackage{multirow}

\begin{document}

\title[Subject-Diffusion]{Subject-Diffusion: Open Domain Personalized Text-to-Image Generation without Test-time Fine-tuning}


\author{Jian Ma}
\orcid{0009-0004-0057-3033}
\affiliation{%
  \institution{OPPO AI Center}
  \city{ShenZhen}
  \country{China}
}
\email{majian2@oppo.com}

\author{Junhao Liang}
\authornote{Work done during internship at OPPO}
\orcid{0000-0001-5612-7631}
\affiliation{%
  \institution{Southern University of Science and Technology}
  \city{ShenZhen}
  \country{China}
}
\email{12132342@mail.sustech.edu.cn}

\author{Chen Chen}
\authornote{corresponding authors}
\orcid{0000-0003-3498-2527}
\affiliation{%
  \institution{OPPO AI Center}
  \city{ShenZhen}
  \country{China}
}
\email{chenchen4@oppo.com}

\author{Haonan Lu}
\authornotemark[2]
\orcid{0000-0001-6332-2785}
\affiliation{%
  \institution{OPPO AI Center}
  \city{ShenZhen}
  \country{China}
}
\email{luhaonan@oppo.com}

\renewcommand{\shortauthors}{Jian Ma, Junhao Liang, Chen Chen, and Haonan Lu.}

\begin{abstract}
Recent progress in personalized image generation using diffusion models has been significant. However, development in the area of open-domain and test-time fine-tuning-free personalized image generation is proceeding rather slowly. In this paper, we propose Subject-Diffusion, a novel open-domain personalized image generation model that, in addition to not requiring test-time fine-tuning, also only requires a single reference image to support personalized generation of single- or two-subjects in any domain. Firstly, we construct an automatic data labeling tool and use the LAION-Aesthetics dataset to construct a large-scale dataset consisting of 76M images and their corresponding subject detection bounding boxes, segmentation masks, and text descriptions. Secondly, we design a new unified framework that combines text and image semantics by incorporating coarse location and fine-grained reference image control to maximize subject fidelity and generalization. Furthermore, we also adopt an attention control mechanism to support two-subject generation. Extensive qualitative and quantitative results demonstrate that our method have certain advantages over other frameworks in single, multiple, and human-customized image generation.
\end{abstract}




\begin{teaserfigure}
	\centering
    \includegraphics[width=0.8\textwidth]{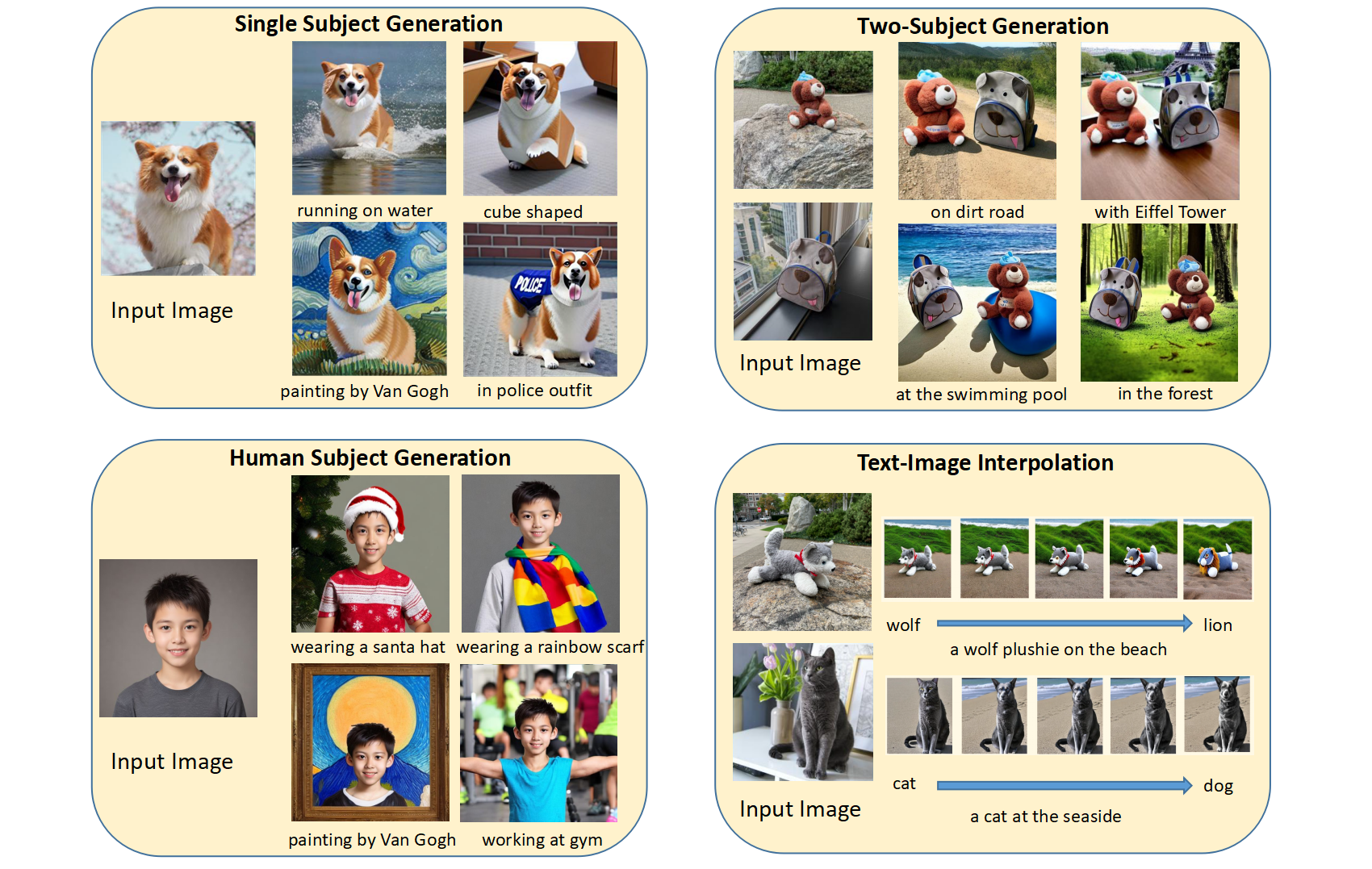}
     \caption{\small{Our Subject-Diffusion is capable of generating high-fidelity subject-driven images (general and human subjects) using just one reference image, without test-time fine-tuning, allowing for the preservation of identity and editability. Furthermore, our model supports the generation of multiple subjects within a single model. We also show the interpolation ability between reference images and word concepts. The original images, except for the left-below boy, are from the DreamBench~\cite{ruiz2023dreambooth} dataset.}}
  \label{fig:banner}
\end{teaserfigure}


\maketitle
\section{Introduction}
Recently, the rapid advancement of diffusion-based generative models~\cite{ho2020denoising,song2020score,song2020denoising} has led to the development of numerous large-scale synthesis models~\cite{rombach2022high,ramesh2022hierarchical,nichol2022glide,saharia2022photorealistic,balaji2022ediffi,feng2023ernie}. These models have been trained on extensive datasets containing billions of image-text pairs, such as LAION-5B~\cite{schuhmann2022laion}, and have demonstrated remarkable text-to-image generation capabilities, exhibiting impressive artistry, authenticity, and semantic alignment. However, relying solely on textual information proves insufficient to fully capture user intentions. Consequently, the integration of textual descriptions and reference images to produce customized images has emerged as a promising research direction.

Leveraging pre-trained text-to-image generation model, such as Stable Diffusion~\cite{rombach2022high} and Imagen~\cite{saharia2022photorealistic}, numerous approaches~\cite{gal2022image,ruiz2023dreambooth,kumari2023multi,tewel2023key,avrahami2023break,hao2023vico,smith2023continual}  propose fine-tuning these models, typically with 3 to 5 provided reference images.
These methods, which often yield satisfactory outcomes, necessitate the execution of specialized network training, as seen in word embedding space~\cite{gal2022image} or in specific layers of the UNet~\cite{ruiz2023dreambooth,kumari2023multi}, or by appending side branches~\cite{smith2023continual}. However, this approach is not efficient for realistic application.
An alternate strategic roadmap~\cite{xiao2023fastcomposer,wei2023elite,chen2023subject,chen2022re,zhou2023customization} is to re-train the text-to-image base model by specially designed network structures or training strategies on a large-scale personalized image dataset. Nonetheless, this often compromises the fidelity and generalization when contrasted with test-time fine-tuning approaches. Additionally, some techniques are only able to produce personalized image generation within specific domains, such as portrait~\cite{xiao2023fastcomposer,shi2023instantbooth, jia2023taming}, cats~\cite{shi2023instantbooth} or dogs~\cite{jia2023taming}. Even though some recent proposed algorithms~\cite{wei2023elite, li2023blip, ma2023unified} can achieve open-domain customized image generation, they can only handle single-concept issues. However, with regard to a single reference image, multiple concept generation, test-time fine-tuning free, and open-domain zero-shot capability, very few papers have considered the comprehensive capability.  

A large-scale training dataset, incorporating object-level segmentation masks and image-level intricate language descriptions, is essential for zero-shot personalized image generation. However, in the case of such labor-intensive labeling tasks, the publicly accessible datasets, including LVIS~\cite{gupta2019lvis}, ADE20K~\cite{zhou2019semantic}, COCO-stuff~\cite{caesar2018coco}, Visual Genome~\cite{krishna2017visual}, and Open Images~\cite{kuznetsova2020open}, usually suffer from insufficient quantities of images ranging from 10k to 1M, or a lack of text descriptions entirely. Addressing this data scarcity issue for open-domain personalized image generation, we are inspired to develop an automatic data labeling tool, which will be thoroughly discussed in Sec.~\ref{sec:dataset}.



As mentioned in~\cite{zhou2023enhancing}, the information of personalized images may overwhelmingly dominate that of the text input of the user to prevent creative generation. In order to balance fidelity and editability, we propose to fuse the input text prompt and object-level image features by continually training the CLIP text encoder (unlike fixing the encoder as FastComposer~\cite{xiao2023fastcomposer} and Elite~\cite{wei2023elite}) based on a specific prompt style. We further propose to integrate fine-grained reference image patches, detected object bounding boxes, and location masks to control the fidelity of generated images. Finally, to further control the generation of multiple subjects, we introduce cross-attention map control during training. As exhibited in Fig.~\ref{fig:banner}, based on the constructed large-scale structured data in an open domain and our proposed new model architecture, Subject-Diffusion achieves remarkable fidelity and editability, which can perform single, multiple, and human subject personalized generation by modifying their shape, pose, background, and even style with only one reference image for each subject. In addition, Subject-Diffusion can also perform smooth interpolation between customized images and text descriptions by using a specially designed denoising process. In terms of quantitative comparisons, our model has certain advantages over other methods, including test-time fine-tuning and non-fine-tuning approaches on the DreamBench~\cite{ruiz2023dreambooth} and our proposed larger open-domain test dataset.


We summarize our contributions in the following aspects: \textbf{(i)} We design an automatic dataset construction pipeline and create a sizable and structured training dataset that comprises 76M open-domain images and 222M entities. \textbf{(ii)} To the best of our knowledge, we propose a personalized image generation framework, which is the first work to address the challenge of simultaneously generating open-domain single- and two-concept personalized images without test-time fine-tuning, solely relying on a single reference image for each subject. \textbf{(iii)} Both quantitative and qualitative experimental results demonstrate the excellent performance of our framework as compared with other methods.

\section{Related Work}

\subsection{Text-to-Image Generation}
The diffusion model has emerged as a promising direction to generate images with high fidelity and diversity based on provided textual input. GLIDE~\cite{nichol2022glide} utilizes an unclassified guide to introduce text conditions into the diffusion process. DALL-E2~\cite{ramesh2022hierarchical} uses a diffusion prior module and cascading diffusion decoders to generate high-resolution images based on the CLIP text encoder. Imagen~\cite{saharia2022photorealistic} emphasizes language understanding and suggests using a large T5 language model to better represent semantics. Latent diffusion model ~\cite{rombach2022high} uses an autoencoder to project images into the latent space and applies the diffusion process to generate latent-level feature maps. Stable diffusion (SD) ~\cite{rombach2022high}, ERNIE-ViLG2.0~\cite{feng2023ernie} and ediffi~\cite{balaji2022ediffi} propose to employ a cross-attention mechanism to inject textual conditions into the diffusion generation process. Our framework is built on the basis of SD due to its flexible scalability and open-source nature.

\subsection{Subject-driven Text-to-Image Generation}
Currently, there are two main frameworks for personalized text-to-image generation from the perspective of whether to introduce test-time fine-tuning. In terms of test-time fine-tuning strategies, a group of solutions requires several personalized images containing a specific subject and then directly fine-tune the token embedding of the subject to adapt to learning visual concepts~\cite{gal2022image,han2023highly,yang2023controllable,voynov2023p+,alaluf2023neural}. Another group of approaches fine-tune the generation model using these images~\cite{ruiz2023dreambooth,kumari2023multi,han2023svdiff,fei2023gradient,chen2023disenbooth}, among which DreamBooth~\cite{ruiz2023dreambooth} fine-tunes the entire UNet network, while Custom Diffusion~\cite{kumari2023multi} only fine-tunes the K and V layers of the cross-attention.
On the other hand, Custom Diffusion proposes the personalized generation of multiple subjects for the first time. SVDiff~\cite{han2023svdiff} constructs training data using cutmix and adds regularization penalties to limit the confusion of multiple subject attention maps. Cones proposes concept neurons and updates only the concept neurons for a single subject in the K and V layers of cross-attention. For multiple personalized subject generation, the concept neurons of multiple trained personalized models are directly concatenated. Mix-of-Show~\cite{gu2023mix} trains a separate LoRA model~\cite{hu2021lora} for each subject and then performs fusion. Cones 2~\cite{liu2023cones2} generates two-subject combination images by learning the residual of token embedding and controlling the attention map.
MagiCapture~\cite{hyung2023magicapture} introduce a multi-concept personalization method capable of generating high-resolution portrait images that faithfully capture the characteristics of both source and reference images.

Since test-time fine-tuning methods suffer from a notoriously time-consuming problem, another research route involves constructing a large amount of domain-specific data or using open-domain image data for training without additional fine-tuning. InstructPix2Pix~\cite{brooks2023instructpix2pix} can follow human instructions to perform various editing tasks, including object swapping, style transfer, and environment modification, by simply concatenating the latent of the reference images during the model noise injection process. ELITE~\cite{wei2023elite} proposes global and local mapping training schemes using the OpenImages testset, which contains 125k images and 600 object classes as training data. However, due to the limitations of the model architecture, the text alignment effect is relatively moderate. UMM-Diffusion presents a novel Unified Multi-Modal Latent Diffusion~\cite{ma2023unified} that takes joint texts and images that contain specified subjects as input sequences and generates customized images with the subjects. Its limitations include the inability to support multiple subjects and its training data being selected from LAION-400M~\cite{schuhmann2021laion}, resulting in poor performance in generating rare themes. Similarly, Taming Encoder~\cite{jia2023taming}, InstantBooth~\cite{shi2023instantbooth}, FastComposer~\cite{xiao2023fastcomposer}, Face-Diffuser~\cite{wang2023highfidelity} and PhotoVerse~\cite{chen2023photoverse} are all trained on domain-specific data. BLIP-Diffusion\cite{li2023blip} uses OpenImages data, and due to its two-stage training scheme, it achieves good fidelity effects but does not support two-subject generation.
IP-Adapter~\cite{ye2023ipadapter} uses a lightweight image prompt adaptation method with the decoupled cross-attention strategy for existing text-to-image diffusion models. SuTI~\cite{chen2023subject} propose a subject-driven text-to-image generator that performs instant and customized generation for a visual subject. Costomization Assistant~\cite{zhou2023customization} employs MLLM to enable test-time tuning-free personalized image generation with more user-friendly interactions.

In contrast, our model, which is trained on a sizable, self-constructed open-domain dataset, performs exceptionally well in terms of the trade-off between fidelity and generalization in both single- and two-subject generation by providing only one reference image for each subject.

\section{Methodology}
\label{sec:methodology}
In this section, we will first introduce our large-scale open-domain dataset for personalized image generation. 
Then, an overview of the Subject-Diffusion framework, followed by an explanation of how we leverage auxiliary information, including reformulating prompts, integrating fine-grained image and location information, and a cross-attention map control strategy, will be detailed.

\subsection{Dataset Construction}
\label{sec:dataset}

In order to equip the diffusion model with the capacity for arbitrary subject image generation, there exists a need for a substantial multimodal dataset that encompasses open-domain capabilities. However, current image datasets either contain a limited number of images, as demonstrated in COCO-Stuff~\cite{caesar2018coco} and OpenImages~\cite{kuznetsova2020open}, or are deficient in modalities such as segmentation masks and detection bounding boxes, coupled with inconsistent data quality as evidenced in LAION-5B~\cite{schuhmann2022laion}. Hence, we are inspired to devise a large-scale, high-quality multimodal dataset that is fitting for our specific task.

\begin{figure*}[h!]
    \begin{centering}
        \setlength{\belowcaptionskip}{-0.2em}
    \setlength{\abovecaptionskip}{0.5em}
      \includegraphics[width=0.65\linewidth]{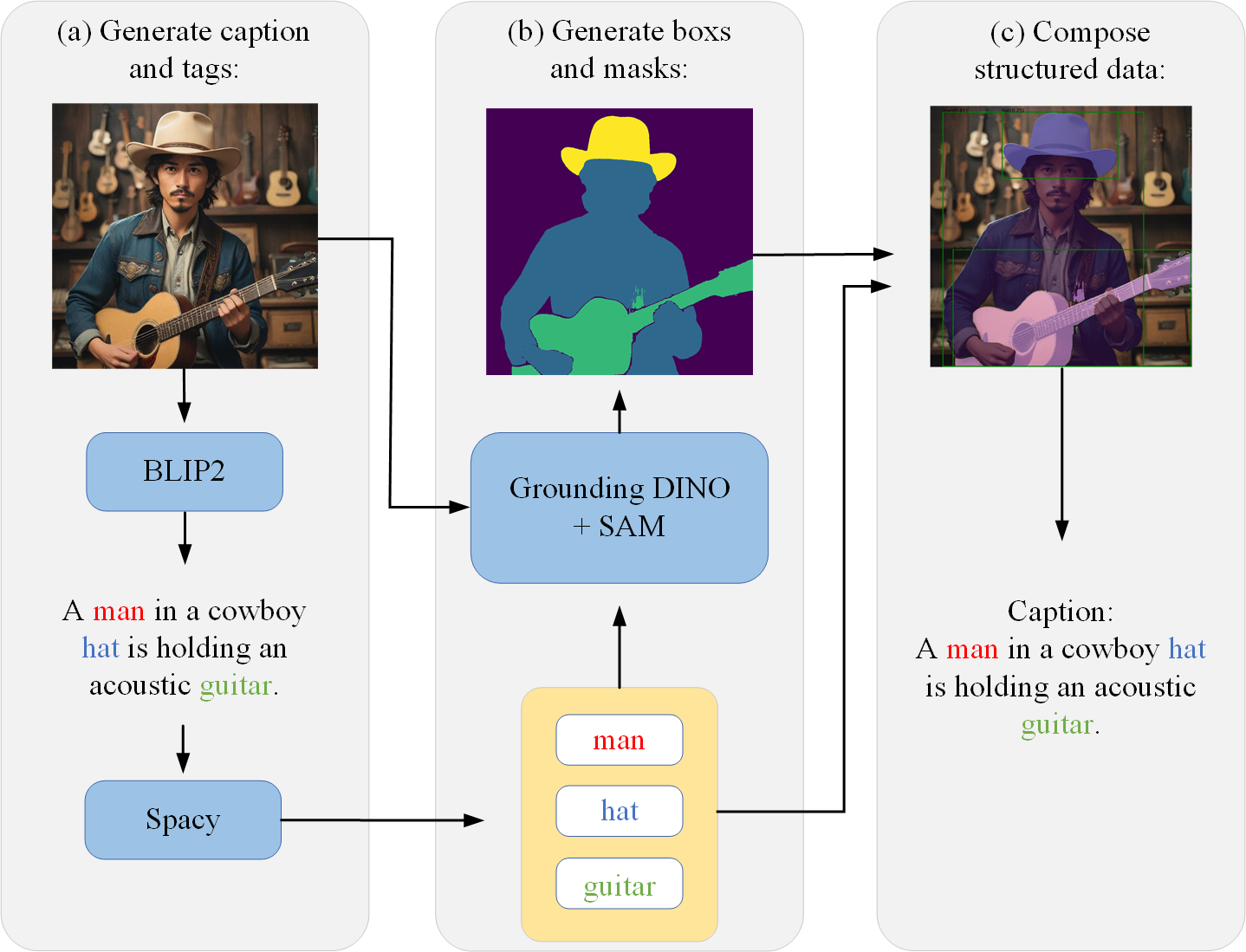}
      \caption{\small{The procedure for training data generation.(a) We first use BLIP2~\cite{li2023blip} to generate the caption of the given image, then we use spacy~\cite{honnibal2020spacy} to extract tags based on the part of speech of each word in the caption sentence. (b) We use the extracted tags as input to Grounding DINO~\cite{liu2023grounding} to obtain detection boxes for each object, and then these detection boxes are used as input prompt to SAM~\cite{kirillov2023segment} to obtain their respective object masks. (c) Finally, all the different modalities are combined into structured data as our multimodal dataset.
      }}
    \label{fig:data_collection}
    \end{centering}
\end{figure*}

As depicted in Fig.~\ref{fig:data_collection}, we outline the three steps we took to create our training data based on LAION-5B. The captions for the images provided by LAION-5B are of poor quality, often containing irrelevant or non-sensical language. This can pose a significant challenge for text-to-image tasks that require accurate image captions. To address this issue, by using BLIP-2~\cite{li2023blip}, we can generate more precise captions for each image.
However, for our subject-driven image generation task, we also need to obtain entities' masks and labels from the images. In order to accomplish this, we perform part-of-speech analysis on the generated captions and treat the nouns as entity tags. Once we have obtained the entity labels, we can use the open-set detection model Grounding DINO~\cite{liu2023grounding} to detect the corresponding location of the entity and use the detection box as a cue for the segmentation model SAM~\cite{kirillov2023segment} to determine the corresponding mask. Finally, we combine the image-text pairs, detection boxes, segmentation masks, and corresponding labels for all instances to structure the data. Based on the aforementioned pipeline, we apply sophisticated filtering strategies, to form the final high-quality dataset called \textbf{S}ubject-\textbf{D}iffusion \textbf{D}ataset (SDD). Our dataset contains 76M examples, 222M entities, and 162K common object classes, which is much larger than the number of annotated images in the famous OpenImages (1M images)~\cite{kuznetsova2020open}. Furthermore, it also covers a wide range of variations involving the capture of scenes, entity classes, and photography conditions (resolution, illumination, \textit{etc.}). This great diversity, as well as its large scale, offers great potential for learning subject-driven image generation abilities in the open domain, which is believed to boost the development of generative artificial intelligence.

\subsection{Model Overview}
\begin{figure*}[tb]
	\centering
    \setlength{\belowcaptionskip}{0.5em}
    \setlength{\abovecaptionskip}{0.5em}
	\includegraphics[width=0.80\textwidth]{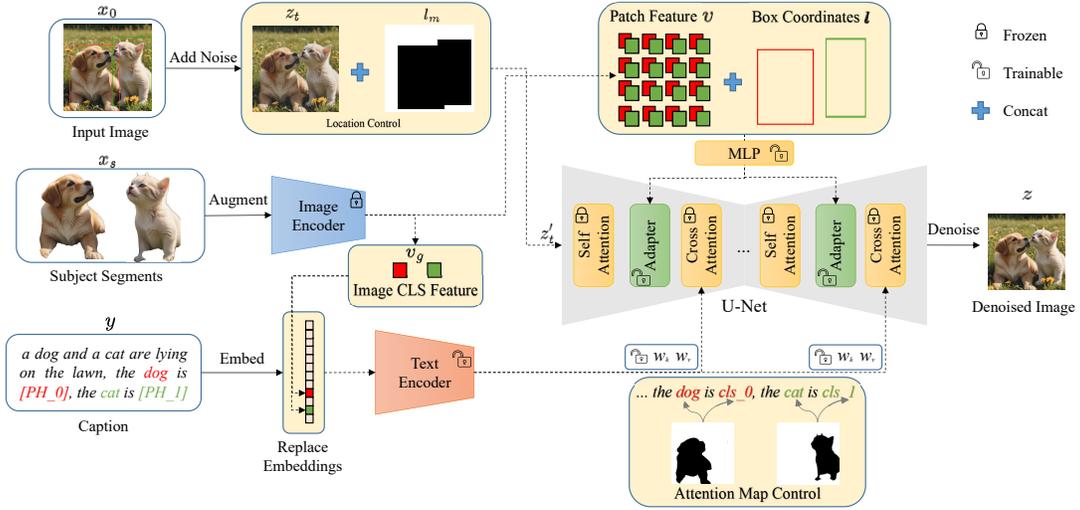}
\caption{\small{An overview of the proposed Subject-Diffusion method based on SD structure. \textbf{(i)} We first design a specific condition by integrating the text prompt and object image features. \textbf{(ii)} Then, we extract fine-grained image local patch features, combining with the detected object bounding boxes, and insert an adapter module between self- and cross-attention in the UNet to enhance the model's fidelity ability. \textbf{(iii)} Further, we propose to employ an attention map control strategy to deal with the multiple object generation issue.}}
\label{fig:Framework}
\end{figure*}

The comprehensive training structure of our suggested Subject-Diffusion is portrayed in Fig.~\ref{fig:Framework}. The design philosophy behind Subject-Diffusion is primarily rooted in three elements. First, we craft a specific prompt format and utilize a text encoder to blend the text with object-level visual features, serving as conditions for SD. Second, to amplify the authenticity of the created personalized images, we suggest integrating an adapter among each self- and cross-attention block. This adapter encodes the dense patch features of the segmented objects along with their corresponding bounding box data. Third, to equip Subject-Diffusion with multiple personalized image generation capabilities, we propose implementing a cross-attention map control strategy, grounded in segmentation masks, to enforce a model that concentrates on local optimization between the entity and its associated area.

\subsection{Exploitation of Auxiliary Information}

\paragraph{Fusion text encoder}
As proposed in Textual Inversion~\cite{gal2022image}, a learned image embedding incorporated with prompt embedding is essential to achieve personalized image generation. Therefore, we first construct a new prompt template similar to BLIP-Diffusion~\cite{li2023blip}: ``\textit{[text prompt], the [subject label 0] is [PH\_0], the [subject label 1] is [PH\_1], ...}'' where ``\textit{text prompt}'' represents the original text description, ``\textit{subject label *}'' represents the category label of the subject, and ``\textit{PH\_*}'' are place holders corresponding to the subject image.
Then, in contrast to approaches~\cite{ma2023glyphdraw,shi2023instantbooth,xiao2023fastcomposer,ma2023unified}, we choose to fuse text and image information before the text encoder. We conduct extensive experiments, showing that fusing text and image information before the text encoder and then retraining the entire text encoder has stronger self-consistency than fusing them later. Specifically, we replace the entity token embedding at the first embedding layer of the text encoder with the image subject ``CLS'' embedding at the corresponding position, and then retrain the entire text encoder.

\paragraph{Dense image and object location control}
Generating personalized images in an open domain while ensuring the fidelity of the subject image with only textual input poses a significant challenge. To address this challenge, we propose to incorporate dense image features as an important input condition, similar to the textual input condition.
To ensure that the model focuses solely on the subject information of the image and disregards the background information, we feed the segmented subject image into the CLIP~\cite{radford2021learning} image encoder to obtain 256-length patch feature tokens. Furthermore, to prevent confusion when generating multiple subjects, we fuse the corresponding image embedding with the Fourier-transformed coordinate position information of the subject. Subsequently, we feed the fused information into the UNet framework for learning, similar to GLIGEN~\cite{li2023gligen}. In each Transformer block, we introduce a new learnable adapter layer between the self-attention layer and the cross-attention layer, which takes the fused information as input and is defined as $\mathcal{L}_a := \mathcal{L}_a + \beta \cdot {tanh(\gamma)} \cdot S([\mathcal{L}_a,h^e])$, 
where $\mathcal{L}_a$ is the output of the self-attention layer, $\beta$ a constant to balance the importance of the adapter layer, $\gamma$ a learnable scalar that is initialized as 0, $S$ the self-attention operator, and $h^e = MLP([v, Fourier(l)])$,
where $MLP(\cdot,\cdot)$ is a multi-layer perceptron that concatenates the two inputs across the feature dimension: $v$ the visual 256 patch feature tokens of an image, and $l$ the coordinate position information of the subject. In the process of training the UNet model, we selectively activate the key and value layers of the cross-attention layers and the adapter layers while freezing the remaining layers. This approach is adopted to enable the model to focus more on learning the adapter layer.

In addition, to prevent model learning from collapsing, a location-area control is innovatively introduced to decouple the distribution between the foreground and background regions. Specifically, as shown in Fig.~\ref{fig:Framework}, a binary mask feature map is generated and concatenated to the original image latent feature for a single subject. For multiple subjects, we overlay the binary images of each subject and then concatenate them onto the latent feature. During inference, the binary image can be specified by the user, detected automatically based on the user's personalized image, or just randomly generated.

\paragraph{Cross attention map control}
Currently, text-to-image generation models often encounter confusion and omissions when generating multiple entities. Most solutions involve controlling the cross-attention map during model inference~\cite{wu2023harnessing,wang2023compositional,rassin2023linguistic,chefer2023attend}.
The approaches proposed in this study are primarily based on the conclusions drawn from Prompt-to-Prompt~\cite{hertz2022prompt,han2023svdiff,xiao2023fastcomposer,avrahami2023break}: The cross-attention in the text-to-image diffusion models can reflect the positions of each generated object specified by the corresponding text token, which is calculated from:
\begin{equation}
{CA}_l(z_t,y_k) = Softmax(Q_l(z_t) \cdot L_l(y_k)^T),
\end{equation}
where ${CA}_l(z_t,y_k)$ is the cross-attention map at layer $l$ of the denoising network between the intermediate feature of the noisy latent $z_t$ and the $k_{th}$ text token $y_k$, and $Q_l$ and $L_l$ are the query and key projections. For each text token, we could get an attention map of size $h_l \times w_l$, where $h_l$ and $w_l$ are the spatial dimensions of the feature $z_t$ and the cross-attention mechanism within diffusion models governs the layout of generated images. The scores in cross-attention maps represent the amount of information that flows from a text token to a latent pixel.
Similarly, we assume that subject confusion arises from an unrestricted cross-attention mechanism, as a single latent pixel can attend to all other tokens.
Therefore, we introduce an additional loss term that encourages the model not only to reconstruct the pixels associated with learned concepts, but also to ensure that each token only attends to the image region occupied by the corresponding concept.
For instance, as illustrated in Fig.~\ref{fig:Framework}, we introduce an attention map regularization term at the position of the entity tokens ``\textit{dog}'' , ``\textit{[cls\_0]}'', ``\textit{cat}'' and ``\textit{[cls\_1]}''. Intuitively, the positions within the area containing the entity
\textit{e.g.,} ``cat'', should have larger values than other positions, so we optimize $z_t$ toward the target that the desired area of the object has large values by penalizing the L1 deviation between the attention maps and the corresponding segmentation maps of the entities. We choose $l$ to be the layers with $h_l=w_l=\{32,16,8\}$.
Formally, we incorporate the following loss terms into the training phase:
\begin{equation}
{L}_{attn} =\frac{1}{N}\sum_{k=1}^N\sum_{l}\left|{CA}_l(z_t,y_k)-M_{k}\right|
\end{equation}
where $M_{k}$ is the segmentation mask of the $k_{th}$ object corresponding to its text token.

\paragraph{Objective function}
As shown in Fig.~\ref{fig:Framework}, given the original clear image $x_0$ and segmented subject image $x_s$, the detected image mask $l_m$ is concatenated to the noisy image latent vector $z_t$ to form a new latent vector $z'_t=concat(z_t, l_m)$. After dimension adjustment through a convolution layer, the feature vector $\Tilde{z}_t=conv\_in(z'_t)$ is fed into the UNet as the query component. In terms of conditional information, given the text prompt $y$, $C=T_\theta(v_g, t_y)$ is fused by the text encoder $T_\theta$ from segmented image global embedding ($v_g = I_\theta(x_s)$) and the text token embeddings ($t_y$) which are extracted from the fixed CLIP image encoder ($I_\theta$) and the text embedding layer, respectively. For the adapters, they receive local image patch features $v$ and bbox coordinates $l$ as additional information through a MLP feature fusion. Consequently, the Subject-Diffusion training objective is:
\begin{equation}
\label{eq:loss}
\mathcal{L} = \mathbb{E}_{\mathcal{E}(x_0),y,\epsilon \sim \mathcal{N}(0,1),t} \big[\parallel\epsilon-\epsilon_\theta(z_t,t,y,x_s,l,l_m)\parallel^2_2\big] + \lambda_{attn} {L}_{attn}.
\end{equation}
where $\lambda_{attn}$ is a weighting hyper-parameter.



\section{Experiments}

\subsection{Implementation Details and Evaluation}

The Subject-Diffusion is trained on SDD, as detailed information is provided in Sec.~\ref{sec:dataset}. We follow the benchmark \textit{DreamBench} proposed in \cite{ruiz2023dreambooth} for quantitative and qualitative comparison. In order to further validate the model's generation capability in the open domain, we also utilize the validation and test data from OpenImages, which comprises 296 classes with two different entity images in each class. In comparison, DreamBench only includes 30 classes.
We evaluate our method with image alignment and text alignment metrics. For image alignment, we calculate the CLIP visual similarity (CLIP-I) and DINO~\cite{caron2021emerging} similarity between the generated images and the target concept images. For text alignment, we calculate the CLIP text-image similarity (CLIP-T) between the generated images and the given text prompts.

We compare several methods for personalized image generation, including Textual Inversion~\cite{gal2022image}, DreamBooth~\cite{ruiz2023dreambooth} and Custom Diffusion~\cite{kumari2023multi}. All of these models require test-time fine-tuning on personalized images in a certain category. Additionally, we compare ELITE~\cite{wei2023elite} and BLIP-Diffusion~\cite{li2023blip}, both are trained on OpenImages without test-time fine-tuning. Another approach compared, IP-Adapter~\cite{ye2023ipadapter}, is trained on a large-scale open-domain dataset without test-time fine-tuning, also.

\subsection{Experiments}


Generating personalized images can be a resource-intensive task, with some methods requiring significant storage and computing power to fine-tune models based on user-provided photos. However, our method and similar ones do not require any test-time fine-tuning and can generate personalized images in a zero-shot manner, making them more efficient and user-friendly. In the following sections, we will present both quantitative and qualitative results of our method as compared with other approaches in both single- and two-subject settings.

\paragraph{Comparison results for single-subject}
We compare our Subject-Diffusion with the 6 aforementioned methods for single-subject generation. In Table~\ref{tab:quatan}, we follow DreamBooth and Blip-Diffusion to generate 6 images for each text prompt provided by DreamBench, amounting in total to 4,500 images for all the subjects. We report the average DINO, CLIP-I, and CLIP-T scores over all pairs of real and generated images. The overall results show that our method significantly outperforms other methods in terms of DINO score, with a score of 0.711 compared to DreamBooth's score of 0.668. Our CLIP-I and CLIP-T scores are also slightly higher or on par with other fine-tuning free algorithms, ELITE and BLIP-Diffusion. However, the CLIP-I score of IP-Adapter is higher than our method (0.813 vs 0.782). Furthermore, we conduct experiments on the OpenImages testset, which has about $10\times$ the number of subjects as DreamBench, and our method still achieve high DINO (0.668), CLIP-I (0.782), and CLIP-T (0.303) scores, revealing its generalization ability.


Fig.~\ref{fig:quali-single} displays a comparison of the qualitative results of single-subject image generation across various prompts, using different approaches. Excluding Textual Inversion, ELITE and IP-adapter, which exhibit significantly lower subject fidelity, our proposed method's subject fidelity and text consistency are comparable to DreamBooth and CustomDiffusion methods that require multiple images for fine-tuning.


\begin{table}[htbp]
\centering
\caption{\small{Quantitative single subject results. DB denotes DreamBench, and OIT represents the OpenImage testset. $\dagger$ indicates experimental results referenced from BLIP-Diffusion. The value of ELITE is tested by ourself. Boldface indicates the best results of zero shot approaches evaluated in Dreambench.All the comparison methods here are based on the SD model.}}
\label{tab:quatan}
\resizebox{1.0\linewidth}{!}{
\begin{tabular}{cccccc}
\toprule
\textbf{Methods}              & \textbf{Type}   & \textbf{Testset}   & \textbf{DINO}& \textbf{CLIP-I} & \textbf{CLIP-T} \\\midrule
Real Images $\dagger$ & -      & -   & 0.774                & 0.885   & -     \\\midrule
Textual Inversion $\dagger$    & FT  & DB & 0.569                & 0.780  & 0.255  \\
DreamBooth $\dagger$          & FT  &DB& 0.668                & 0.803  &0.305  \\
Custom Diffusion    & FT  & DB & 0.643           & 0.790  & 0.305 \\\midrule
ELITE       & ZS &DB& 0.621  & 0.771  & 0.293  \\
BLIP-Diffusion $\dagger$       & ZS &DB& 0.594                & 0.779  & \textbf{0.300}  \\
IP-Adapter $\dagger$       & ZS &DB& 0.667          & \textbf{0.813}  & 0.289 
\\\midrule
\multirow{2}{*}{\textbf{Subject-Diffusion}}&\multirow{2}{*}{ZS} &DB&  \textbf{0.711} & 0.787  & 0.293 \\
&  & OIT& 0.668 & 0.782  &  0.303 \\\bottomrule
\end{tabular}}
\end{table}

\paragraph{Comparison result for two-subject}

We conduct a comparison study on our method with two fine-tuning-based approaches, \textit{i.e.,} DreamBooth and Custom Diffusion. This study involves 30 different combinations of two subjects from DreamBench. For each combination, we generated 6 images per prompt by utilizing 25 text prompts from DreamBench.
As depicted in Fig.~\ref{fig:quali-multi}, we present five prompts of generated images. Overall, our method demonstrates superior performance compared to the other two methods, particularly in maintaining subject fidelity in the generated images. On the one hand, images generated by the comparative methods often miss one subject, as exemplified by DreamBooth's failure to include entities like ``on cobblestone street'' and ``floating on water'', as well as Custom Diffusion's inability to accurately capture entities in ``on dirty road'' and ``on cobblestone street''. On the other hand, while these methods are capable of generating two subjects, the appearance features between them are noticeably leaking and mixing, leading to lower subject fidelity when compared to the images provided by the user.
By contrast, the images generated by our method effectively preserve the user-provided subjects, and each one is accurately produced.

We also calculate DINO, CLIP-I and CLIP-T scores on all groups of the generated images, user-provided images and prompts. To obtain CLIP-I, we average the calculated similarities between the generated image and the two subjects, as results presented in Table~\ref{tab:multi_quatan}. Obviously, our approach shows remarkable superiority over DreamBooth and Custom Diffusion across DINO and CLIP-T, providing compelling evidence of its ability to capture the subject information of reference images more accurately and display multiple entities in a single image simultaneously.

\begin{table}[htbp]
\centering
\caption{\small{Quantitative result of two subject generation. ZS means zero-shot and FT denotes fine-tuning. Boldface indicates the best results.}}
\label{tab:multi_quatan}
\resizebox{1\linewidth}{!}{
\begin{tabular}{ccccc}
\toprule
\textbf{Methods} & \textbf{Type} & \textbf{DINO} & \textbf{CLIP-I} & \textbf{CLIP-T} \\\midrule
DreamBooth & FT & 0.430 &  0.695 &0.308  \\
Custom Diffusion & FT & 0.464  & \textbf{0.698} & 0.300  \\
\textbf{Subject-Diffusion} & ZS & \textbf{0.506} & 0.696 & \textbf{0.310} \\\bottomrule
\end{tabular}}
\end{table}

\subsection{Ablation Studies}
\label{sec:ablation}
The ablation studies involve examining two main aspects, namely: 1) the impact of our training data and 2) the impact of different components in our Subject-Diffusion model.
As shown in Table~\ref{tab:abl}, we present zero-shot evaluation results for both single- and two-subject cases. We observe that all the ablation settings result in weaker quantitative results than our full setting.

\begin{table}[htbp]
\centering
\caption{\small{Ablation Results.$\uparrow$ and $\downarrow$ indicate increase or decrease, respectively. Boldface indicates full-setting results.}}
\label{tab:abl}
\resizebox{1\linewidth}{!}{
\begin{tabular}{cccccc}
\toprule
 & Index & \textbf{Methods} & \textbf{DINO} & \textbf{CLIP-I} & \textbf{CLIP-T} \\\midrule
 &(a) &  \textbf{Subject-Diffusion} & \textbf{0.711} & \textbf{0.787} & \textbf{0.293} \\
 &(b)  &trained on OpenImage&   0.664$\downarrow$ & 0.777$\downarrow$  & 0.294$\uparrow$ \\
  \textbf{Single} &(c)&  w/o location control  &0.694$\downarrow$ & 0.778$\downarrow$ & 0.275$\downarrow$  \\
   &(d)&  w/o box coordinates  & 0.732$\uparrow$ & 0.810$\uparrow$ & 0.282$\downarrow$  \\
 \textbf{Subject} &(e)&  w/o adapter layer & 0.534$\downarrow$ & 0.731$\downarrow$ & 0.291$\downarrow$  \\
&(f) &  w/o attention map control  & 0.692$\downarrow$ & 0.789$\uparrow$ & 0.288$\downarrow $  \\
 &(g) &  w/o image cls feature & 0.637$\downarrow$& 0.719$\downarrow$ & 0.299$\uparrow$  \\
   \midrule
 &(a) & \textbf{Subject-Diffusion}  & \textbf{0.506} & \textbf{0.696} & \textbf{0.310} \\
 &(b) &trained on OpenImage&   0.491$\downarrow$ & 0.693$\downarrow$  & 0.302$\downarrow$ \\
  \textbf{Two}&(c)&  w/o location control  & 0.477$\downarrow$& 0.666$\downarrow$  & 0.281$\downarrow$  \\
   &(d) &  w/o box coordinates  & 0.464$\downarrow$ & 0.687$\downarrow$ & 0.305$\downarrow$  \\
 \textbf{Subjects}&(e)&  w/o adapter layer & 0.411$\downarrow$  & 0.649$\downarrow$ & 0.307$\downarrow$  \\

&(f)&  w/o attention map control  & 0.500$\downarrow$ & 0.688$\downarrow$ & 0.302$\downarrow$  \\
&(g) &  w/o image cls feature & 0.457$\downarrow$ & 0.627$\downarrow$ & 0.309$\downarrow$  \\
\bottomrule
\end{tabular}}
\end{table}
\paragraph{Impact of our training data}
The training data proposed in this paper consists of large-scale, richly annotated images, thereby enabling our model to effectively capture the appearance features of any given subject. To further assess the impact of training data, we retrain our model using OpenImages~\cite{kuznetsova2020open} training data, limiting the categories to only 600. Our evaluation results (a) and (b) demonstrate that this smaller dataset leads to lower image similarity, with the DINO and CLIP-I scores both decreasing for single-subject and two-subject cases, which underscores the importance of utilizing large-scale training data in generating highly personalized images. However, the results still surpass or are on par with those of ELITE and BLIP-diffusion (0.664 vs. 0.621 vs. 0.594 for DINO), demonstrating the effectiveness of Subject-Diffusion's model structure and training strategy.

\paragraph{Impact of different components}
The comparison between experiments (a) and (c) declares that, if we remove the \textit{location control} (object masks), our model will apparently degenerate over all evaluation metrics.
Experiments (a) and (d) indicate that the introduction of \textit{box coordinates} leads to significant improvements in two-subject generation (with the DINO score increasing by 0.042, the CLIP-I score increasing by 0.09, and the CLIP-T score increasing by 0.005). However, the fidelity of single-subject generation decreased by 0.021 for the DINO score and 0.023 for the CLIP-I score.
This decline may be due to the fact that, when generating a single subject, the information becomes overly redundant, making it challenging for the model to grasp the key details of the subject.

The high fidelity of our model is primarily attributed to the 256 image patch features input to the adapter layer. As demonstrated in experiment (e), removing this module results in a significant drop in nearly all of the metrics.
Experimental results (f) clearly indicate that the \textit{attention map control} operation delivers a substantial performance improvement for two-subject generation as well as a slight performance improvement for single-subject generation. This difference is most likely due to the ability of the attention map control mechanism to prevent confusion between different subjects.
The results of (a) and (g) indicate that the absence of the image ``CLS'' feature led to a significant reduction in the fidelity of the subject, highlighting the significance of the feature in representing the overall image information.

\subsection{Human Image Generation}

\begin{table}[htbp]
\centering
\caption{\small{Comparison among our method and baselines on single-subject human image generation. $\dagger$ indicates that the experimental values are referenced from FastComposer.}}
\label{tab:human}
\begin{tabular}{cccc}
\toprule
Method                    & Images$\downarrow$ & ID Preser.$\uparrow$ & Prompt Consis.$\uparrow$ \\
\midrule
StableDiffusion$\dagger$           & 0      & 0.039                 & 0.268              \\
\midrule
Textual-Inversion$\dagger$         & 5      & 0.293                 & 0.219              \\
DreamBooth$\dagger$                & 5      & 0.273                 & 0.239              \\
Custom Diffusion$\dagger$          & 5      & 0.434                 & 0.233              \\
FastComposer$\dagger$              & 1      & 0.514                 & \textbf{0.243}       \\
IP-Adapter$\dagger$              & 1      & 0.520                 & 0.201       \\
\textbf{Subject-Diffusion} & \textbf{1}      & \textbf{0.605}                 & 0.228 \\
\bottomrule
\end{tabular}
\end{table}
Due to our method's ability to produce high-fidelity results, it is also well-suited for human image generation. To evaluate our model's effectiveness in this area, we use the single-entity evaluation method employed in FastComposer~\cite{xiao2023fastcomposer} and compare our model's performance to that of other existing methods.
The experimental results are shown in Table~\ref{tab:human}. Subject-Diffusion significantly outperforms all baseline approaches in identity preservation, with an exceptionally high score that surpasses FastComposer trained on the specific portrait dataset by 0.091. However, in terms of prompt consistency, our method is slightly weaker than FastComposer (-0.015). We believe this vulnerability could be due to our method's tendency to prioritize subject fidelity when dealing with challenging prompt words.

\subsection{Text-Image Interpolation}

By utilizing the ``[text prompt], the [subject label] is [PH]'' prompt template during image generation, we are able to utilize the dual semantics of both text and image to control the generated image output. Moreover, we could utilize texts and images from distinct categories and perform interpolation of generated images by controlling the proportion of the diffusion steps.
To achieve this, we remove the user input image control once the image layout is generated, retaining only the textual semantic control. Our step-based interpolation method is represented by the following formula:
\begin{equation}
    \epsilon_t =
    \begin{cases}
        \epsilon_\theta(z_t, t, y',x_s,l,l_m)  & \text{if } t > \alpha T, \\
        \epsilon_\theta(z_t, t, y) & \text{otherwise}
    \end{cases}
    \\
\end{equation}
In this context, $y$ denotes the original text prompt, while $y'$ signifies employing a convoluted text template: "[text prompt]$_a$, the [subject]$_b$ is [cls]$_a$". 
The visualization examples can be found in Fig.~\ref{fig:interpolation}. We provide this experiment to show that the high-level information of the user-provided images are successfully extracted and rendered in generated images during early backward diffusion stages. Thus we can adjust $\alpha$ to balance image fidelity and editablity according to different prompts.

\section{User Study}
We perform a user study to analyze the fidelity and prompt consistency of the generated images. we follow DreamBooth and BLIP-Diffusion to generate 6 images for each text prompt provided by DreamBench, amounting in total to 4,500 images for all the subjects. Considering whether the models or demos of the comparison methods are open-sourced, we mainly compared our method with three methods, ELITE, IP-Adapter and BLIP-Diffusion. Our scoring rules are as follows: for each image, annotator independently score it based on two criteria - the fidelity of the generated image and prompt consistency. Scores range from 1 to 5 points. As the score increases, the fidelity and prompt consistency become stronger. The final results of the User study are shown in Table.~\ref{tab:user_study}. The results of each method will ultimately be averaged.

Our method has a slight advantage in prompt consistency metric compared to BLIP-Diffusion, and has a significant advantage in the fidelity metric compared to all three methods. ELITE has a significant advantage in prompt consistency, but sacrifices image fidelity, while IP-Adapter achieves a better balance between the two metrics.
Simultaneously, the differences between the annotation results and parts of the conclusions in Table~\ref{tab:quatan} indicate that objective metrics cannot truly reflect human preferences.
The annotation results further demonstrate the mutual constraint between these two metrics, so simultaneously improving them is an important future research direction for personalized image generation.

\begin{table}[htbp]
\centering
\caption{\small{Qualitative single subject results with user study.}}
\label{tab:user_study}
\begin{tabular}{ccc}
\toprule
Scores    &   ID Preser.$\uparrow$ & Prompt Consis.$\uparrow$ \\
\midrule
ELITE         & 1.7928               & \textbf{2.5276}          \\
IP-Adapter            & 2.2178               & 2.3213             \\
BLIP-Diffusion            & 1.9330                & 2.2293           \\
\textbf{Subject-Diffusion}        & \textbf{3.4748}        & 2.2689            \\
\bottomrule
\end{tabular}
\end{table}

\section{Conclusion and Limitation}
Thus far, the considerable cost and limited availability of manual labeling have presented profound challenges to the pragmatic deployment of personalized image generation models. Drawing inspiration from advancements in zero-shot large models, this study introduces an automated data labeling tool to assemble a large-scale structured image dataset. Subsequently, we establish a comprehensive framework that merges text and image semantics, leveraging various tiers of information to optimize subject fidelity and generalization. Empirical analysis from our experiments demonstrate that our methodology surpasses existing models on the DreamBench dataset, suggesting potential to serve as a foundation for enhancing the efficiency of personalized image generation models within the open domain.

\paragraph{Limitation}
Although our method is capable of zero-shot generation with any reference image in open domains and can handle multi-subject scenarios, it still has certain limitations. First, our method faces challenges in editing attributes and accessories within user-input images, leading to limitations in the scope of the model's applicability. Secondly, when generating personalized images for more than two subjects, our model will fail to render harmonious images with a high probability. 
In the future, we will conduct further research to address these shortcomings.

\clearpage

\bibliographystyle{ACM-Reference-Format}
\bibliography{subject}


\begin{thebibliography}{60}


\ifx \showCODEN    \undefined \def \showCODEN     #1{\unskip}     \fi
\ifx \showDOI      \undefined \def \showDOI       #1{#1}\fi
\ifx \showISBNx    \undefined \def \showISBNx     #1{\unskip}     \fi
\ifx \showISBNxiii \undefined \def \showISBNxiii  #1{\unskip}     \fi
\ifx \showISSN     \undefined \def \showISSN      #1{\unskip}     \fi
\ifx \showLCCN     \undefined \def \showLCCN      #1{\unskip}     \fi
\ifx \shownote     \undefined \def \shownote      #1{#1}          \fi
\ifx \showarticletitle \undefined \def \showarticletitle #1{#1}   \fi
\ifx \showURL      \undefined \def \showURL       {\relax}        \fi
\providecommand\bibfield[2]{#2}
\providecommand\bibinfo[2]{#2}
\providecommand\natexlab[1]{#1}
\providecommand\showeprint[2][]{arXiv:#2}

\bibitem[Alaluf et~al\mbox{.}(2023)]%
        {alaluf2023neural}
\bibfield{author}{\bibinfo{person}{Yuval Alaluf}, \bibinfo{person}{Elad Richardson}, \bibinfo{person}{Gal Metzer}, {and} \bibinfo{person}{Daniel Cohen-Or}.} \bibinfo{year}{2023}\natexlab{}.
\newblock \showarticletitle{A Neural Space-Time Representation for Text-to-Image Personalization}.
\newblock \bibinfo{journal}{\emph{arXiv preprint arXiv:2305.15391}} (\bibinfo{year}{2023}).
\newblock


\bibitem[Avrahami et~al\mbox{.}(2023)]%
        {avrahami2023break}
\bibfield{author}{\bibinfo{person}{Omri Avrahami}, \bibinfo{person}{Kfir Aberman}, \bibinfo{person}{Ohad Fried}, \bibinfo{person}{Daniel Cohen-Or}, {and} \bibinfo{person}{Dani Lischinski}.} \bibinfo{year}{2023}\natexlab{}.
\newblock \showarticletitle{Break-A-Scene: Extracting Multiple Concepts from a Single Image}.
\newblock \bibinfo{journal}{\emph{arXiv preprint arXiv:2305.16311}} (\bibinfo{year}{2023}).
\newblock


\bibitem[Balaji et~al\mbox{.}(2022)]%
        {balaji2022ediffi}
\bibfield{author}{\bibinfo{person}{Yogesh Balaji}, \bibinfo{person}{Seungjun Nah}, \bibinfo{person}{Xun Huang}, \bibinfo{person}{Arash Vahdat}, \bibinfo{person}{Jiaming Song}, \bibinfo{person}{Karsten Kreis}, \bibinfo{person}{Miika Aittala}, \bibinfo{person}{Timo Aila}, \bibinfo{person}{Samuli Laine}, \bibinfo{person}{Bryan Catanzaro}, {et~al\mbox{.}}} \bibinfo{year}{2022}\natexlab{}.
\newblock \showarticletitle{ediffi: Text-to-image diffusion models with an ensemble of expert denoisers}.
\newblock \bibinfo{journal}{\emph{arXiv preprint arXiv:2211.01324}} (\bibinfo{year}{2022}).
\newblock


\bibitem[Brooks et~al\mbox{.}(2023)]%
        {brooks2023instructpix2pix}
\bibfield{author}{\bibinfo{person}{Tim Brooks}, \bibinfo{person}{Aleksander Holynski}, {and} \bibinfo{person}{Alexei~A Efros}.} \bibinfo{year}{2023}\natexlab{}.
\newblock \showarticletitle{Instructpix2pix: Learning to follow image editing instructions}. In \bibinfo{booktitle}{\emph{Proceedings of the IEEE/CVF Conference on Computer Vision and Pattern Recognition}}. \bibinfo{pages}{18392--18402}.
\newblock


\bibitem[Caesar et~al\mbox{.}(2018)]%
        {caesar2018coco}
\bibfield{author}{\bibinfo{person}{Holger Caesar}, \bibinfo{person}{Jasper Uijlings}, {and} \bibinfo{person}{Vittorio Ferrari}.} \bibinfo{year}{2018}\natexlab{}.
\newblock \showarticletitle{Coco-stuff: Thing and stuff classes in context}. In \bibinfo{booktitle}{\emph{Proceedings of the IEEE conference on computer vision and pattern recognition}}. \bibinfo{pages}{1209--1218}.
\newblock


\bibitem[Caron et~al\mbox{.}(2021)]%
        {caron2021emerging}
\bibfield{author}{\bibinfo{person}{Mathilde Caron}, \bibinfo{person}{Hugo Touvron}, \bibinfo{person}{Ishan Misra}, \bibinfo{person}{Herv{\'e} J{\'e}gou}, \bibinfo{person}{Julien Mairal}, \bibinfo{person}{Piotr Bojanowski}, {and} \bibinfo{person}{Armand Joulin}.} \bibinfo{year}{2021}\natexlab{}.
\newblock \showarticletitle{Emerging properties in self-supervised vision transformers}. In \bibinfo{booktitle}{\emph{Proceedings of the IEEE/CVF international conference on computer vision}}. \bibinfo{pages}{9650--9660}.
\newblock


\bibitem[Chefer et~al\mbox{.}(2023)]%
        {chefer2023attend}
\bibfield{author}{\bibinfo{person}{Hila Chefer}, \bibinfo{person}{Yuval Alaluf}, \bibinfo{person}{Yael Vinker}, \bibinfo{person}{Lior Wolf}, {and} \bibinfo{person}{Daniel Cohen-Or}.} \bibinfo{year}{2023}\natexlab{}.
\newblock \showarticletitle{Attend-and-excite: Attention-based semantic guidance for text-to-image diffusion models}.
\newblock \bibinfo{journal}{\emph{arXiv preprint arXiv:2301.13826}} (\bibinfo{year}{2023}).
\newblock


\bibitem[Chen et~al\mbox{.}(2023b)]%
        {chen2023disenbooth}
\bibfield{author}{\bibinfo{person}{Hong Chen}, \bibinfo{person}{Yipeng Zhang}, \bibinfo{person}{Xin Wang}, \bibinfo{person}{Xuguang Duan}, \bibinfo{person}{Yuwei Zhou}, {and} \bibinfo{person}{Wenwu Zhu}.} \bibinfo{year}{2023}\natexlab{b}.
\newblock \showarticletitle{Disenbooth: Identity-preserving disentangled tuning for subject-driven text-to-image generation}.
\newblock \bibinfo{journal}{\emph{arXiv preprint arXiv:2305.03374}} (\bibinfo{year}{2023}).
\newblock


\bibitem[Chen et~al\mbox{.}(2023c)]%
        {chen2023photoverse}
\bibfield{author}{\bibinfo{person}{Li Chen}, \bibinfo{person}{Mengyi Zhao}, \bibinfo{person}{Yiheng Liu}, \bibinfo{person}{Mingxu Ding}, \bibinfo{person}{Yangyang Song}, \bibinfo{person}{Shizun Wang}, \bibinfo{person}{Xu Wang}, \bibinfo{person}{Hao Yang}, \bibinfo{person}{Jing Liu}, \bibinfo{person}{Kang Du}, {et~al\mbox{.}}} \bibinfo{year}{2023}\natexlab{c}.
\newblock \showarticletitle{PhotoVerse: Tuning-Free Image Customization with Text-to-Image Diffusion Models}.
\newblock \bibinfo{journal}{\emph{arXiv preprint arXiv:2309.05793}} (\bibinfo{year}{2023}).
\newblock


\bibitem[Chen et~al\mbox{.}(2023a)]%
        {chen2023subject}
\bibfield{author}{\bibinfo{person}{Wenhu Chen}, \bibinfo{person}{Hexiang Hu}, \bibinfo{person}{Yandong Li}, \bibinfo{person}{Nataniel Rui}, \bibinfo{person}{Xuhui Jia}, \bibinfo{person}{Ming-Wei Chang}, {and} \bibinfo{person}{William~W Cohen}.} \bibinfo{year}{2023}\natexlab{a}.
\newblock \showarticletitle{Subject-driven text-to-image generation via apprenticeship learning}.
\newblock \bibinfo{journal}{\emph{arXiv preprint arXiv:2304.00186}} (\bibinfo{year}{2023}).
\newblock


\bibitem[Chen et~al\mbox{.}(2022)]%
        {chen2022re}
\bibfield{author}{\bibinfo{person}{Wenhu Chen}, \bibinfo{person}{Hexiang Hu}, \bibinfo{person}{Chitwan Saharia}, {and} \bibinfo{person}{William~W Cohen}.} \bibinfo{year}{2022}\natexlab{}.
\newblock \showarticletitle{Re-imagen: Retrieval-augmented text-to-image generator}.
\newblock \bibinfo{journal}{\emph{arXiv preprint arXiv:2209.14491}} (\bibinfo{year}{2022}).
\newblock


\bibitem[Fei et~al\mbox{.}(2023)]%
        {fei2023gradient}
\bibfield{author}{\bibinfo{person}{Zhengcong Fei}, \bibinfo{person}{Mingyuan Fan}, {and} \bibinfo{person}{Junshi Huang}.} \bibinfo{year}{2023}\natexlab{}.
\newblock \showarticletitle{Gradient-Free Textual Inversion}.
\newblock \bibinfo{journal}{\emph{arXiv preprint arXiv:2304.05818}} (\bibinfo{year}{2023}).
\newblock


\bibitem[Feng et~al\mbox{.}(2023)]%
        {feng2023ernie}
\bibfield{author}{\bibinfo{person}{Zhida Feng}, \bibinfo{person}{Zhenyu Zhang}, \bibinfo{person}{Xintong Yu}, \bibinfo{person}{Yewei Fang}, \bibinfo{person}{Lanxin Li}, \bibinfo{person}{Xuyi Chen}, \bibinfo{person}{Yuxiang Lu}, \bibinfo{person}{Jiaxiang Liu}, \bibinfo{person}{Weichong Yin}, \bibinfo{person}{Shikun Feng}, {et~al\mbox{.}}} \bibinfo{year}{2023}\natexlab{}.
\newblock \showarticletitle{ERNIE-ViLG 2.0: Improving text-to-image diffusion model with knowledge-enhanced mixture-of-denoising-experts}. In \bibinfo{booktitle}{\emph{Proceedings of the IEEE/CVF Conference on Computer Vision and Pattern Recognition}}. \bibinfo{pages}{10135--10145}.
\newblock


\bibitem[Gal et~al\mbox{.}(2022)]%
        {gal2022image}
\bibfield{author}{\bibinfo{person}{Rinon Gal}, \bibinfo{person}{Yuval Alaluf}, \bibinfo{person}{Yuval Atzmon}, \bibinfo{person}{Or Patashnik}, \bibinfo{person}{Amit~H Bermano}, \bibinfo{person}{Gal Chechik}, {and} \bibinfo{person}{Daniel Cohen-Or}.} \bibinfo{year}{2022}\natexlab{}.
\newblock \showarticletitle{An image is worth one word: Personalizing text-to-image generation using textual inversion}.
\newblock \bibinfo{journal}{\emph{arXiv preprint arXiv:2208.01618}} (\bibinfo{year}{2022}).
\newblock


\bibitem[Gu et~al\mbox{.}(2023)]%
        {gu2023mix}
\bibfield{author}{\bibinfo{person}{Yuchao Gu}, \bibinfo{person}{Xintao Wang}, \bibinfo{person}{Jay~Zhangjie Wu}, \bibinfo{person}{Yujun Shi}, \bibinfo{person}{Yunpeng Chen}, \bibinfo{person}{Zihan Fan}, \bibinfo{person}{Wuyou Xiao}, \bibinfo{person}{Rui Zhao}, \bibinfo{person}{Shuning Chang}, \bibinfo{person}{Weijia Wu}, {et~al\mbox{.}}} \bibinfo{year}{2023}\natexlab{}.
\newblock \showarticletitle{Mix-of-Show: Decentralized Low-Rank Adaptation for Multi-Concept Customization of Diffusion Models}.
\newblock \bibinfo{journal}{\emph{arXiv preprint arXiv:2305.18292}} (\bibinfo{year}{2023}).
\newblock


\bibitem[Gupta et~al\mbox{.}(2019)]%
        {gupta2019lvis}
\bibfield{author}{\bibinfo{person}{Agrim Gupta}, \bibinfo{person}{Piotr Dollar}, {and} \bibinfo{person}{Ross Girshick}.} \bibinfo{year}{2019}\natexlab{}.
\newblock \showarticletitle{Lvis: A dataset for large vocabulary instance segmentation}. In \bibinfo{booktitle}{\emph{Proceedings of the IEEE/CVF conference on computer vision and pattern recognition}}. \bibinfo{pages}{5356--5364}.
\newblock


\bibitem[Han et~al\mbox{.}(2023b)]%
        {han2023highly}
\bibfield{author}{\bibinfo{person}{Inhwa Han}, \bibinfo{person}{Serin Yang}, \bibinfo{person}{Taesung Kwon}, {and} \bibinfo{person}{Jong~Chul Ye}.} \bibinfo{year}{2023}\natexlab{b}.
\newblock \showarticletitle{Highly Personalized Text Embedding for Image Manipulation by Stable Diffusion}.
\newblock \bibinfo{journal}{\emph{arXiv preprint arXiv:2303.08767}} (\bibinfo{year}{2023}).
\newblock


\bibitem[Han et~al\mbox{.}(2023a)]%
        {han2023svdiff}
\bibfield{author}{\bibinfo{person}{Ligong Han}, \bibinfo{person}{Yinxiao Li}, \bibinfo{person}{Han Zhang}, \bibinfo{person}{Peyman Milanfar}, \bibinfo{person}{Dimitris Metaxas}, {and} \bibinfo{person}{Feng Yang}.} \bibinfo{year}{2023}\natexlab{a}.
\newblock \showarticletitle{Svdiff: Compact parameter space for diffusion fine-tuning}.
\newblock \bibinfo{journal}{\emph{arXiv preprint arXiv:2303.11305}} (\bibinfo{year}{2023}).
\newblock


\bibitem[Hao et~al\mbox{.}(2023)]%
        {hao2023vico}
\bibfield{author}{\bibinfo{person}{Shaozhe Hao}, \bibinfo{person}{Kai Han}, \bibinfo{person}{Shihao Zhao}, {and} \bibinfo{person}{Kwan-Yee~K Wong}.} \bibinfo{year}{2023}\natexlab{}.
\newblock \showarticletitle{ViCo: Detail-Preserving Visual Condition for Personalized Text-to-Image Generation}.
\newblock \bibinfo{journal}{\emph{arXiv preprint arXiv:2306.00971}} (\bibinfo{year}{2023}).
\newblock


\bibitem[Hertz et~al\mbox{.}(2022)]%
        {hertz2022prompt}
\bibfield{author}{\bibinfo{person}{Amir Hertz}, \bibinfo{person}{Ron Mokady}, \bibinfo{person}{Jay Tenenbaum}, \bibinfo{person}{Kfir Aberman}, \bibinfo{person}{Yael Pritch}, {and} \bibinfo{person}{Daniel Cohen-Or}.} \bibinfo{year}{2022}\natexlab{}.
\newblock \showarticletitle{Prompt-to-prompt image editing with cross attention control}.
\newblock \bibinfo{journal}{\emph{arXiv preprint arXiv:2208.01626}} (\bibinfo{year}{2022}).
\newblock


\bibitem[Ho et~al\mbox{.}(2020)]%
        {ho2020denoising}
\bibfield{author}{\bibinfo{person}{Jonathan Ho}, \bibinfo{person}{Ajay Jain}, {and} \bibinfo{person}{Pieter Abbeel}.} \bibinfo{year}{2020}\natexlab{}.
\newblock \showarticletitle{Denoising diffusion probabilistic models}.
\newblock \bibinfo{journal}{\emph{Advances in Neural Information Processing Systems}}  \bibinfo{volume}{33} (\bibinfo{year}{2020}), \bibinfo{pages}{6840--6851}.
\newblock


\bibitem[Honnibal et~al\mbox{.}(2020)]%
        {honnibal2020spacy}
\bibfield{author}{\bibinfo{person}{Matthew Honnibal}, \bibinfo{person}{Ines Montani}, \bibinfo{person}{Sofie Van~Landeghem}, \bibinfo{person}{Adriane Boyd}, {et~al\mbox{.}}} \bibinfo{year}{2020}\natexlab{}.
\newblock \showarticletitle{spaCy: Industrial-strength natural language processing in python}.
\newblock  (\bibinfo{year}{2020}).
\newblock


\bibitem[Hu et~al\mbox{.}(2021)]%
        {hu2021lora}
\bibfield{author}{\bibinfo{person}{Edward~J Hu}, \bibinfo{person}{Yelong Shen}, \bibinfo{person}{Phillip Wallis}, \bibinfo{person}{Zeyuan Allen-Zhu}, \bibinfo{person}{Yuanzhi Li}, \bibinfo{person}{Shean Wang}, \bibinfo{person}{Lu Wang}, {and} \bibinfo{person}{Weizhu Chen}.} \bibinfo{year}{2021}\natexlab{}.
\newblock \showarticletitle{Lora: Low-rank adaptation of large language models}.
\newblock \bibinfo{journal}{\emph{arXiv preprint arXiv:2106.09685}} (\bibinfo{year}{2021}).
\newblock


\bibitem[Hyung et~al\mbox{.}(2023)]%
        {hyung2023magicapture}
\bibfield{author}{\bibinfo{person}{Junha Hyung}, \bibinfo{person}{Jaeyo Shin}, {and} \bibinfo{person}{Jaegul Choo}.} \bibinfo{year}{2023}\natexlab{}.
\newblock \bibinfo{title}{MagiCapture: High-Resolution Multi-Concept Portrait Customization}.
\newblock
\newblock
\showeprint[arxiv]{2309.06895}~[cs.CV]


\bibitem[Jia et~al\mbox{.}(2023)]%
        {jia2023taming}
\bibfield{author}{\bibinfo{person}{Xuhui Jia}, \bibinfo{person}{Yang Zhao}, \bibinfo{person}{Kelvin~CK Chan}, \bibinfo{person}{Yandong Li}, \bibinfo{person}{Han Zhang}, \bibinfo{person}{Boqing Gong}, \bibinfo{person}{Tingbo Hou}, \bibinfo{person}{Huisheng Wang}, {and} \bibinfo{person}{Yu-Chuan Su}.} \bibinfo{year}{2023}\natexlab{}.
\newblock \showarticletitle{Taming encoder for zero fine-tuning image customization with text-to-image diffusion models}.
\newblock \bibinfo{journal}{\emph{arXiv preprint arXiv:2304.02642}} (\bibinfo{year}{2023}).
\newblock


\bibitem[Kirillov et~al\mbox{.}(2023)]%
        {kirillov2023segment}
\bibfield{author}{\bibinfo{person}{Alexander Kirillov}, \bibinfo{person}{Eric Mintun}, \bibinfo{person}{Nikhila Ravi}, \bibinfo{person}{Hanzi Mao}, \bibinfo{person}{Chloe Rolland}, \bibinfo{person}{Laura Gustafson}, \bibinfo{person}{Tete Xiao}, \bibinfo{person}{Spencer Whitehead}, \bibinfo{person}{Alexander~C Berg}, \bibinfo{person}{Wan-Yen Lo}, {et~al\mbox{.}}} \bibinfo{year}{2023}\natexlab{}.
\newblock \showarticletitle{Segment anything}.
\newblock \bibinfo{journal}{\emph{arXiv preprint arXiv:2304.02643}} (\bibinfo{year}{2023}).
\newblock


\bibitem[Krishna et~al\mbox{.}(2017)]%
        {krishna2017visual}
\bibfield{author}{\bibinfo{person}{Ranjay Krishna}, \bibinfo{person}{Yuke Zhu}, \bibinfo{person}{Oliver Groth}, \bibinfo{person}{Justin Johnson}, \bibinfo{person}{Kenji Hata}, \bibinfo{person}{Joshua Kravitz}, \bibinfo{person}{Stephanie Chen}, \bibinfo{person}{Yannis Kalantidis}, \bibinfo{person}{Li-Jia Li}, \bibinfo{person}{David~A Shamma}, {et~al\mbox{.}}} \bibinfo{year}{2017}\natexlab{}.
\newblock \showarticletitle{Visual genome: Connecting language and vision using crowdsourced dense image annotations}.
\newblock \bibinfo{journal}{\emph{International Journal of Computer Vision}}  \bibinfo{volume}{123} (\bibinfo{year}{2017}), \bibinfo{pages}{32--73}.
\newblock


\bibitem[Kumari et~al\mbox{.}(2023)]%
        {kumari2023multi}
\bibfield{author}{\bibinfo{person}{Nupur Kumari}, \bibinfo{person}{Bingliang Zhang}, \bibinfo{person}{Richard Zhang}, \bibinfo{person}{Eli Shechtman}, {and} \bibinfo{person}{Jun-Yan Zhu}.} \bibinfo{year}{2023}\natexlab{}.
\newblock \showarticletitle{Multi-concept customization of text-to-image diffusion}. In \bibinfo{booktitle}{\emph{Proceedings of the IEEE/CVF Conference on Computer Vision and Pattern Recognition}}. \bibinfo{pages}{1931--1941}.
\newblock


\bibitem[Kuznetsova et~al\mbox{.}(2020)]%
        {kuznetsova2020open}
\bibfield{author}{\bibinfo{person}{Alina Kuznetsova}, \bibinfo{person}{Hassan Rom}, \bibinfo{person}{Neil Alldrin}, \bibinfo{person}{Jasper Uijlings}, \bibinfo{person}{Ivan Krasin}, \bibinfo{person}{Jordi Pont-Tuset}, \bibinfo{person}{Shahab Kamali}, \bibinfo{person}{Stefan Popov}, \bibinfo{person}{Matteo Malloci}, \bibinfo{person}{Alexander Kolesnikov}, {et~al\mbox{.}}} \bibinfo{year}{2020}\natexlab{}.
\newblock \showarticletitle{The open images dataset v4: Unified image classification, object detection, and visual relationship detection at scale}.
\newblock \bibinfo{journal}{\emph{International Journal of Computer Vision}} \bibinfo{volume}{128}, \bibinfo{number}{7} (\bibinfo{year}{2020}), \bibinfo{pages}{1956--1981}.
\newblock


\bibitem[Li et~al\mbox{.}(2023a)]%
        {li2023blip}
\bibfield{author}{\bibinfo{person}{Dongxu Li}, \bibinfo{person}{Junnan Li}, {and} \bibinfo{person}{Steven~CH Hoi}.} \bibinfo{year}{2023}\natexlab{a}.
\newblock \showarticletitle{Blip-diffusion: Pre-trained subject representation for controllable text-to-image generation and editing}.
\newblock \bibinfo{journal}{\emph{arXiv preprint arXiv:2305.14720}} (\bibinfo{year}{2023}).
\newblock


\bibitem[Li et~al\mbox{.}(2023b)]%
        {li2023gligen}
\bibfield{author}{\bibinfo{person}{Yuheng Li}, \bibinfo{person}{Haotian Liu}, \bibinfo{person}{Qingyang Wu}, \bibinfo{person}{Fangzhou Mu}, \bibinfo{person}{Jianwei Yang}, \bibinfo{person}{Jianfeng Gao}, \bibinfo{person}{Chunyuan Li}, {and} \bibinfo{person}{Yong~Jae Lee}.} \bibinfo{year}{2023}\natexlab{b}.
\newblock \showarticletitle{Gligen: Open-set grounded text-to-image generation}. In \bibinfo{booktitle}{\emph{Proceedings of the IEEE/CVF Conference on Computer Vision and Pattern Recognition}}. \bibinfo{pages}{22511--22521}.
\newblock


\bibitem[Liu et~al\mbox{.}(2023a)]%
        {liu2023grounding}
\bibfield{author}{\bibinfo{person}{Shilong Liu}, \bibinfo{person}{Zhaoyang Zeng}, \bibinfo{person}{Tianhe Ren}, \bibinfo{person}{Feng Li}, \bibinfo{person}{Hao Zhang}, \bibinfo{person}{Jie Yang}, \bibinfo{person}{Chunyuan Li}, \bibinfo{person}{Jianwei Yang}, \bibinfo{person}{Hang Su}, \bibinfo{person}{Jun Zhu}, {et~al\mbox{.}}} \bibinfo{year}{2023}\natexlab{a}.
\newblock \showarticletitle{Grounding dino: Marrying dino with grounded pre-training for open-set object detection}.
\newblock \bibinfo{journal}{\emph{arXiv preprint arXiv:2303.05499}} (\bibinfo{year}{2023}).
\newblock


\bibitem[Liu et~al\mbox{.}(2023b)]%
        {liu2023cones2}
\bibfield{author}{\bibinfo{person}{Zhiheng Liu}, \bibinfo{person}{Yifei Zhang}, \bibinfo{person}{Yujun Shen}, \bibinfo{person}{Kecheng Zheng}, \bibinfo{person}{Kai Zhu}, \bibinfo{person}{Ruili Feng}, \bibinfo{person}{Yu Liu}, \bibinfo{person}{Deli Zhao}, \bibinfo{person}{Jingren Zhou}, {and} \bibinfo{person}{Yang Cao}.} \bibinfo{year}{2023}\natexlab{b}.
\newblock \showarticletitle{Cones 2: Customizable Image Synthesis with Multiple Subjects}.
\newblock \bibinfo{journal}{\emph{arXiv preprint arXiv:2305.19327}} (\bibinfo{year}{2023}).
\newblock


\bibitem[Ma et~al\mbox{.}(2023b)]%
        {ma2023glyphdraw}
\bibfield{author}{\bibinfo{person}{Jian Ma}, \bibinfo{person}{Mingjun Zhao}, \bibinfo{person}{Chen Chen}, \bibinfo{person}{Ruichen Wang}, \bibinfo{person}{Di Niu}, \bibinfo{person}{Haonan Lu}, {and} \bibinfo{person}{Xiaodong Lin}.} \bibinfo{year}{2023}\natexlab{b}.
\newblock \showarticletitle{GlyphDraw: Learning to Draw Chinese Characters in Image Synthesis Models Coherently}.
\newblock \bibinfo{journal}{\emph{arXiv preprint arXiv:2303.17870}} (\bibinfo{year}{2023}).
\newblock


\bibitem[Ma et~al\mbox{.}(2023a)]%
        {ma2023unified}
\bibfield{author}{\bibinfo{person}{Yiyang Ma}, \bibinfo{person}{Huan Yang}, \bibinfo{person}{Wenjing Wang}, \bibinfo{person}{Jianlong Fu}, {and} \bibinfo{person}{Jiaying Liu}.} \bibinfo{year}{2023}\natexlab{a}.
\newblock \showarticletitle{Unified multi-modal latent diffusion for joint subject and text conditional image generation}.
\newblock \bibinfo{journal}{\emph{arXiv preprint arXiv:2303.09319}} (\bibinfo{year}{2023}).
\newblock


\bibitem[Nichol et~al\mbox{.}(2022)]%
        {nichol2022glide}
\bibfield{author}{\bibinfo{person}{Alexander~Quinn Nichol}, \bibinfo{person}{Prafulla Dhariwal}, \bibinfo{person}{Aditya Ramesh}, \bibinfo{person}{Pranav Shyam}, \bibinfo{person}{Pamela Mishkin}, \bibinfo{person}{Bob Mcgrew}, \bibinfo{person}{Ilya Sutskever}, {and} \bibinfo{person}{Mark Chen}.} \bibinfo{year}{2022}\natexlab{}.
\newblock \showarticletitle{GLIDE: Towards Photorealistic Image Generation and Editing with Text-Guided Diffusion Models}. In \bibinfo{booktitle}{\emph{International Conference on Machine Learning}}. PMLR, \bibinfo{pages}{16784--16804}.
\newblock


\bibitem[Radford et~al\mbox{.}(2021)]%
        {radford2021learning}
\bibfield{author}{\bibinfo{person}{Alec Radford}, \bibinfo{person}{Jong~Wook Kim}, \bibinfo{person}{Chris Hallacy}, \bibinfo{person}{Aditya Ramesh}, \bibinfo{person}{Gabriel Goh}, \bibinfo{person}{Sandhini Agarwal}, \bibinfo{person}{Girish Sastry}, \bibinfo{person}{Amanda Askell}, \bibinfo{person}{Pamela Mishkin}, \bibinfo{person}{Jack Clark}, {et~al\mbox{.}}} \bibinfo{year}{2021}\natexlab{}.
\newblock \showarticletitle{Learning transferable visual models from natural language supervision}. In \bibinfo{booktitle}{\emph{International conference on machine learning}}. PMLR, \bibinfo{pages}{8748--8763}.
\newblock


\bibitem[Ramesh et~al\mbox{.}(2022)]%
        {ramesh2022hierarchical}
\bibfield{author}{\bibinfo{person}{Aditya Ramesh}, \bibinfo{person}{Prafulla Dhariwal}, \bibinfo{person}{Alex Nichol}, \bibinfo{person}{Casey Chu}, {and} \bibinfo{person}{Mark Chen}.} \bibinfo{year}{2022}\natexlab{}.
\newblock \showarticletitle{Hierarchical text-conditional image generation with clip latents}.
\newblock \bibinfo{journal}{\emph{arXiv preprint arXiv:2204.06125}} (\bibinfo{year}{2022}).
\newblock


\bibitem[Rassin et~al\mbox{.}(2023)]%
        {rassin2023linguistic}
\bibfield{author}{\bibinfo{person}{Royi Rassin}, \bibinfo{person}{Eran Hirsch}, \bibinfo{person}{Daniel Glickman}, \bibinfo{person}{Shauli Ravfogel}, \bibinfo{person}{Yoav Goldberg}, {and} \bibinfo{person}{Gal Chechik}.} \bibinfo{year}{2023}\natexlab{}.
\newblock \showarticletitle{Linguistic Binding in Diffusion Models: Enhancing Attribute Correspondence through Attention Map Alignment}.
\newblock \bibinfo{journal}{\emph{arXiv preprint arXiv:2306.08877}} (\bibinfo{year}{2023}).
\newblock


\bibitem[Rombach et~al\mbox{.}(2022)]%
        {rombach2022high}
\bibfield{author}{\bibinfo{person}{Robin Rombach}, \bibinfo{person}{Andreas Blattmann}, \bibinfo{person}{Dominik Lorenz}, \bibinfo{person}{Patrick Esser}, {and} \bibinfo{person}{Bj{\"o}rn Ommer}.} \bibinfo{year}{2022}\natexlab{}.
\newblock \showarticletitle{High-resolution image synthesis with latent diffusion models}. In \bibinfo{booktitle}{\emph{Proceedings of the IEEE/CVF Conference on Computer Vision and Pattern Recognition}}. \bibinfo{pages}{10684--10695}.
\newblock


\bibitem[Ruiz et~al\mbox{.}(2023)]%
        {ruiz2023dreambooth}
\bibfield{author}{\bibinfo{person}{Nataniel Ruiz}, \bibinfo{person}{Yuanzhen Li}, \bibinfo{person}{Varun Jampani}, \bibinfo{person}{Yael Pritch}, \bibinfo{person}{Michael Rubinstein}, {and} \bibinfo{person}{Kfir Aberman}.} \bibinfo{year}{2023}\natexlab{}.
\newblock \showarticletitle{Dreambooth: Fine tuning text-to-image diffusion models for subject-driven generation}. In \bibinfo{booktitle}{\emph{Proceedings of the IEEE/CVF Conference on Computer Vision and Pattern Recognition}}. \bibinfo{pages}{22500--22510}.
\newblock


\bibitem[Saharia et~al\mbox{.}(2022)]%
        {saharia2022photorealistic}
\bibfield{author}{\bibinfo{person}{Chitwan Saharia}, \bibinfo{person}{William Chan}, \bibinfo{person}{Saurabh Saxena}, \bibinfo{person}{Lala Li}, \bibinfo{person}{Jay Whang}, \bibinfo{person}{Emily~L Denton}, \bibinfo{person}{Kamyar Ghasemipour}, \bibinfo{person}{Raphael Gontijo~Lopes}, \bibinfo{person}{Burcu Karagol~Ayan}, \bibinfo{person}{Tim Salimans}, {et~al\mbox{.}}} \bibinfo{year}{2022}\natexlab{}.
\newblock \showarticletitle{Photorealistic text-to-image diffusion models with deep language understanding}.
\newblock \bibinfo{journal}{\emph{Advances in Neural Information Processing Systems}}  \bibinfo{volume}{35} (\bibinfo{year}{2022}), \bibinfo{pages}{36479--36494}.
\newblock


\bibitem[Schuhmann et~al\mbox{.}(2022)]%
        {schuhmann2022laion}
\bibfield{author}{\bibinfo{person}{Christoph Schuhmann}, \bibinfo{person}{Romain Beaumont}, \bibinfo{person}{Richard Vencu}, \bibinfo{person}{Cade Gordon}, \bibinfo{person}{Ross Wightman}, \bibinfo{person}{Mehdi Cherti}, \bibinfo{person}{Theo Coombes}, \bibinfo{person}{Aarush Katta}, \bibinfo{person}{Clayton Mullis}, \bibinfo{person}{Mitchell Wortsman}, {et~al\mbox{.}}} \bibinfo{year}{2022}\natexlab{}.
\newblock \showarticletitle{Laion-5b: An open large-scale dataset for training next generation image-text models}.
\newblock \bibinfo{journal}{\emph{arXiv preprint arXiv:2210.08402}} (\bibinfo{year}{2022}).
\newblock


\bibitem[Schuhmann et~al\mbox{.}(2021)]%
        {schuhmann2021laion}
\bibfield{author}{\bibinfo{person}{Christoph Schuhmann}, \bibinfo{person}{Richard Vencu}, \bibinfo{person}{Romain Beaumont}, \bibinfo{person}{Robert Kaczmarczyk}, \bibinfo{person}{Clayton Mullis}, \bibinfo{person}{Aarush Katta}, \bibinfo{person}{Theo Coombes}, \bibinfo{person}{Jenia Jitsev}, {and} \bibinfo{person}{Aran Komatsuzaki}.} \bibinfo{year}{2021}\natexlab{}.
\newblock \showarticletitle{Laion-400m: Open dataset of clip-filtered 400 million image-text pairs}.
\newblock \bibinfo{journal}{\emph{arXiv preprint arXiv:2111.02114}} (\bibinfo{year}{2021}).
\newblock


\bibitem[Shi et~al\mbox{.}(2023)]%
        {shi2023instantbooth}
\bibfield{author}{\bibinfo{person}{Jing Shi}, \bibinfo{person}{Wei Xiong}, \bibinfo{person}{Zhe Lin}, {and} \bibinfo{person}{Hyun~Joon Jung}.} \bibinfo{year}{2023}\natexlab{}.
\newblock \showarticletitle{Instantbooth: Personalized text-to-image generation without test-time finetuning}.
\newblock \bibinfo{journal}{\emph{arXiv preprint arXiv:2304.03411}} (\bibinfo{year}{2023}).
\newblock


\bibitem[Smith et~al\mbox{.}(2023)]%
        {smith2023continual}
\bibfield{author}{\bibinfo{person}{James~Seale Smith}, \bibinfo{person}{Yen-Chang Hsu}, \bibinfo{person}{Lingyu Zhang}, \bibinfo{person}{Ting Hua}, \bibinfo{person}{Zsolt Kira}, \bibinfo{person}{Yilin Shen}, {and} \bibinfo{person}{Hongxia Jin}.} \bibinfo{year}{2023}\natexlab{}.
\newblock \showarticletitle{Continual diffusion: Continual customization of text-to-image diffusion with c-lora}.
\newblock \bibinfo{journal}{\emph{arXiv preprint arXiv:2304.06027}} (\bibinfo{year}{2023}).
\newblock


\bibitem[Song et~al\mbox{.}(2020a)]%
        {song2020denoising}
\bibfield{author}{\bibinfo{person}{Jiaming Song}, \bibinfo{person}{Chenlin Meng}, {and} \bibinfo{person}{Stefano Ermon}.} \bibinfo{year}{2020}\natexlab{a}.
\newblock \showarticletitle{Denoising diffusion implicit models}.
\newblock \bibinfo{journal}{\emph{arXiv preprint arXiv:2010.02502}} (\bibinfo{year}{2020}).
\newblock


\bibitem[Song et~al\mbox{.}(2020b)]%
        {song2020score}
\bibfield{author}{\bibinfo{person}{Yang Song}, \bibinfo{person}{Jascha Sohl-Dickstein}, \bibinfo{person}{Diederik~P Kingma}, \bibinfo{person}{Abhishek Kumar}, \bibinfo{person}{Stefano Ermon}, {and} \bibinfo{person}{Ben Poole}.} \bibinfo{year}{2020}\natexlab{b}.
\newblock \showarticletitle{Score-based generative modeling through stochastic differential equations}.
\newblock \bibinfo{journal}{\emph{arXiv preprint arXiv:2011.13456}} (\bibinfo{year}{2020}).
\newblock


\bibitem[Tewel et~al\mbox{.}(2023)]%
        {tewel2023key}
\bibfield{author}{\bibinfo{person}{Yoad Tewel}, \bibinfo{person}{Rinon Gal}, \bibinfo{person}{Gal Chechik}, {and} \bibinfo{person}{Yuval Atzmon}.} \bibinfo{year}{2023}\natexlab{}.
\newblock \showarticletitle{Key-locked rank one editing for text-to-image personalization}.
\newblock \bibinfo{journal}{\emph{arXiv preprint arXiv:2305.01644}} (\bibinfo{year}{2023}).
\newblock


\bibitem[Voynov et~al\mbox{.}(2023)]%
        {voynov2023p+}
\bibfield{author}{\bibinfo{person}{Andrey Voynov}, \bibinfo{person}{Qinghao Chu}, \bibinfo{person}{Daniel Cohen-Or}, {and} \bibinfo{person}{Kfir Aberman}.} \bibinfo{year}{2023}\natexlab{}.
\newblock \showarticletitle{$ P+ $: Extended Textual Conditioning in Text-to-Image Generation}.
\newblock \bibinfo{journal}{\emph{arXiv preprint arXiv:2303.09522}} (\bibinfo{year}{2023}).
\newblock


\bibitem[Wang et~al\mbox{.}(2023a)]%
        {wang2023compositional}
\bibfield{author}{\bibinfo{person}{Ruichen Wang}, \bibinfo{person}{Zekang Chen}, \bibinfo{person}{Chen Chen}, \bibinfo{person}{Jian Ma}, \bibinfo{person}{Haonan Lu}, {and} \bibinfo{person}{Xiaodong Lin}.} \bibinfo{year}{2023}\natexlab{a}.
\newblock \showarticletitle{Compositional text-to-image synthesis with attention map control of diffusion models}.
\newblock \bibinfo{journal}{\emph{arXiv preprint arXiv:2305.13921}} (\bibinfo{year}{2023}).
\newblock


\bibitem[Wang et~al\mbox{.}(2023b)]%
        {wang2023highfidelity}
\bibfield{author}{\bibinfo{person}{Yibin Wang}, \bibinfo{person}{Weizhong Zhang}, \bibinfo{person}{Jianwei Zheng}, {and} \bibinfo{person}{Cheng Jin}.} \bibinfo{year}{2023}\natexlab{b}.
\newblock \bibinfo{title}{High-fidelity Person-centric Subject-to-Image Synthesis}.
\newblock
\newblock
\showeprint[arxiv]{2311.10329}~[cs.CV]


\bibitem[Wei et~al\mbox{.}(2023)]%
        {wei2023elite}
\bibfield{author}{\bibinfo{person}{Yuxiang Wei}, \bibinfo{person}{Yabo Zhang}, \bibinfo{person}{Zhilong Ji}, \bibinfo{person}{Jinfeng Bai}, \bibinfo{person}{Lei Zhang}, {and} \bibinfo{person}{Wangmeng Zuo}.} \bibinfo{year}{2023}\natexlab{}.
\newblock \showarticletitle{Elite: Encoding visual concepts into textual embeddings for customized text-to-image generation}.
\newblock \bibinfo{journal}{\emph{arXiv preprint arXiv:2302.13848}} (\bibinfo{year}{2023}).
\newblock


\bibitem[Wu et~al\mbox{.}(2023)]%
        {wu2023harnessing}
\bibfield{author}{\bibinfo{person}{Qiucheng Wu}, \bibinfo{person}{Yujian Liu}, \bibinfo{person}{Handong Zhao}, \bibinfo{person}{Trung Bui}, \bibinfo{person}{Zhe Lin}, \bibinfo{person}{Yang Zhang}, {and} \bibinfo{person}{Shiyu Chang}.} \bibinfo{year}{2023}\natexlab{}.
\newblock \showarticletitle{Harnessing the spatial-temporal attention of diffusion models for high-fidelity text-to-image synthesis}.
\newblock \bibinfo{journal}{\emph{arXiv preprint arXiv:2304.03869}} (\bibinfo{year}{2023}).
\newblock


\bibitem[Xiao et~al\mbox{.}(2023)]%
        {xiao2023fastcomposer}
\bibfield{author}{\bibinfo{person}{Guangxuan Xiao}, \bibinfo{person}{Tianwei Yin}, \bibinfo{person}{William~T Freeman}, \bibinfo{person}{Fr{\'e}do Durand}, {and} \bibinfo{person}{Song Han}.} \bibinfo{year}{2023}\natexlab{}.
\newblock \showarticletitle{FastComposer: Tuning-Free Multi-Subject Image Generation with Localized Attention}.
\newblock \bibinfo{journal}{\emph{arXiv preprint arXiv:2305.10431}} (\bibinfo{year}{2023}).
\newblock


\bibitem[Yang et~al\mbox{.}(2023)]%
        {yang2023controllable}
\bibfield{author}{\bibinfo{person}{Jianan Yang}, \bibinfo{person}{Haobo Wang}, \bibinfo{person}{Ruixuan Xiao}, \bibinfo{person}{Sai Wu}, \bibinfo{person}{Gang Chen}, {and} \bibinfo{person}{Junbo Zhao}.} \bibinfo{year}{2023}\natexlab{}.
\newblock \showarticletitle{Controllable Textual Inversion for Personalized Text-to-Image Generation}.
\newblock \bibinfo{journal}{\emph{arXiv preprint arXiv:2304.05265}} (\bibinfo{year}{2023}).
\newblock


\bibitem[Ye et~al\mbox{.}(2023)]%
        {ye2023ipadapter}
\bibfield{author}{\bibinfo{person}{Hu Ye}, \bibinfo{person}{Jun Zhang}, \bibinfo{person}{Sibo Liu}, \bibinfo{person}{Xiao Han}, {and} \bibinfo{person}{Wei Yang}.} \bibinfo{year}{2023}\natexlab{}.
\newblock \bibinfo{title}{IP-Adapter: Text Compatible Image Prompt Adapter for Text-to-Image Diffusion Models}.
\newblock
\newblock
\showeprint[arxiv]{2308.06721}~[cs.CV]


\bibitem[Zhou et~al\mbox{.}(2019)]%
        {zhou2019semantic}
\bibfield{author}{\bibinfo{person}{Bolei Zhou}, \bibinfo{person}{Hang Zhao}, \bibinfo{person}{Xavier Puig}, \bibinfo{person}{Tete Xiao}, \bibinfo{person}{Sanja Fidler}, \bibinfo{person}{Adela Barriuso}, {and} \bibinfo{person}{Antonio Torralba}.} \bibinfo{year}{2019}\natexlab{}.
\newblock \showarticletitle{Semantic understanding of scenes through the ade20k dataset}.
\newblock \bibinfo{journal}{\emph{International Journal of Computer Vision}}  \bibinfo{volume}{127} (\bibinfo{year}{2019}), \bibinfo{pages}{302--321}.
\newblock


\bibitem[Zhou et~al\mbox{.}(2023a)]%
        {zhou2023customization}
\bibfield{author}{\bibinfo{person}{Yufan Zhou}, \bibinfo{person}{Ruiyi Zhang}, \bibinfo{person}{Jiuxiang Gu}, {and} \bibinfo{person}{Tong Sun}.} \bibinfo{year}{2023}\natexlab{a}.
\newblock \bibinfo{title}{Customization Assistant for Text-to-image Generation}.
\newblock
\newblock
\showeprint[arxiv]{2312.03045}~[cs.CV]


\bibitem[Zhou et~al\mbox{.}(2023b)]%
        {zhou2023enhancing}
\bibfield{author}{\bibinfo{person}{Yufan Zhou}, \bibinfo{person}{Ruiyi Zhang}, \bibinfo{person}{Tong Sun}, {and} \bibinfo{person}{Jinhui Xu}.} \bibinfo{year}{2023}\natexlab{b}.
\newblock \showarticletitle{Enhancing Detail Preservation for Customized Text-to-Image Generation: A Regularization-Free Approach}.
\newblock \bibinfo{journal}{\emph{arXiv preprint arXiv:2305.13579}} (\bibinfo{year}{2023}).
\newblock


\end{thebibliography}

\appendix

\begin{figure*}[htbp]
	\centering
	\includegraphics[width=0.9\linewidth]{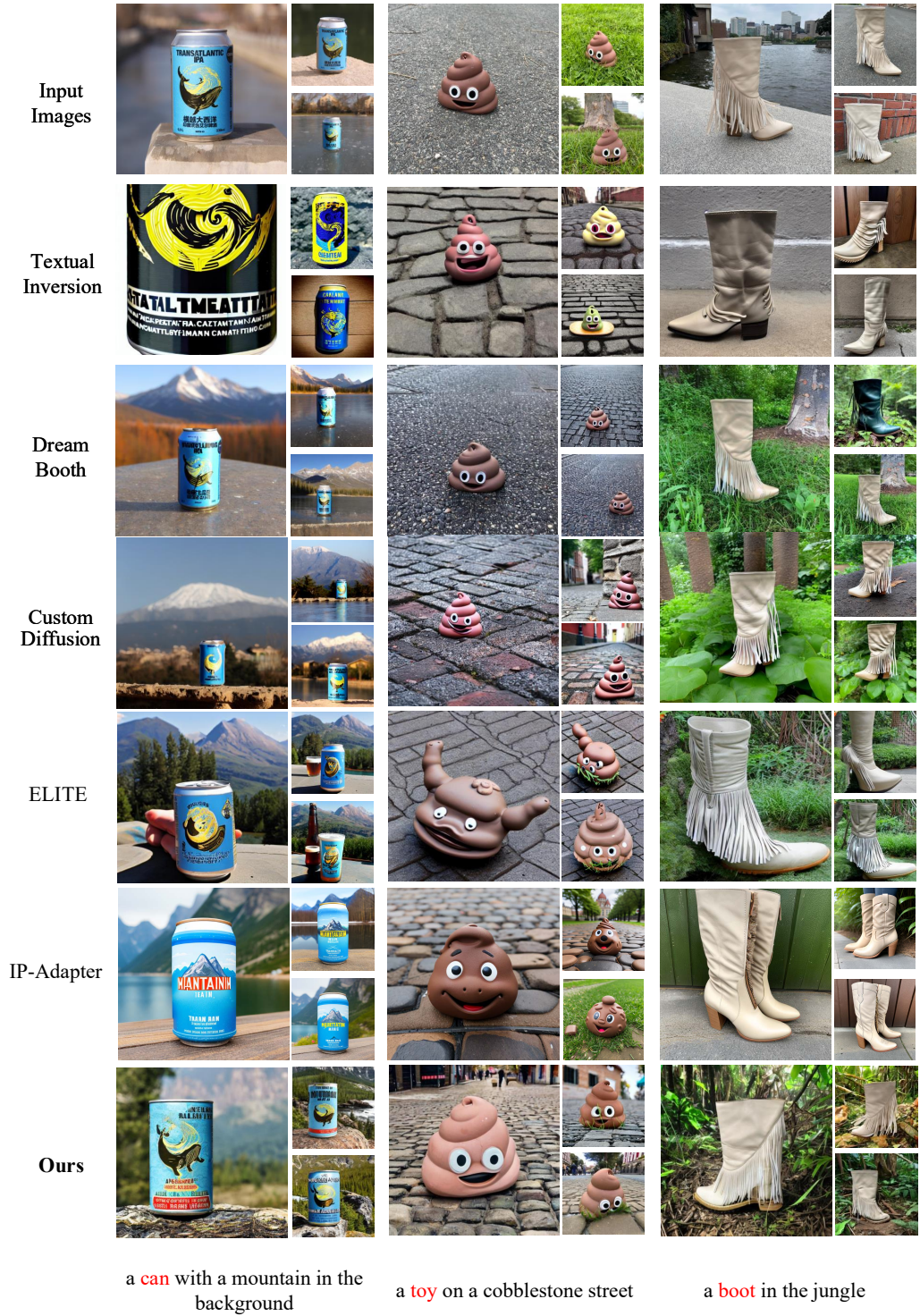}
 \caption{\small{Qualitative result for single-subject generation. Textual Inversion, DreamBooth and Custom Diffusion employ all three reference images to fine-tune models, whereas only ELITE and Subject-Diffusion can generate personalized images using a single input reference image  (corresponding position) without fine-tuning. All original images are from the DreamBench~\cite{ruiz2023dreambooth} dataset.}}
\label{fig:quali-single}
\end{figure*}

\begin{figure*}[htbp]
	\centering
	\includegraphics[width=0.9\linewidth]{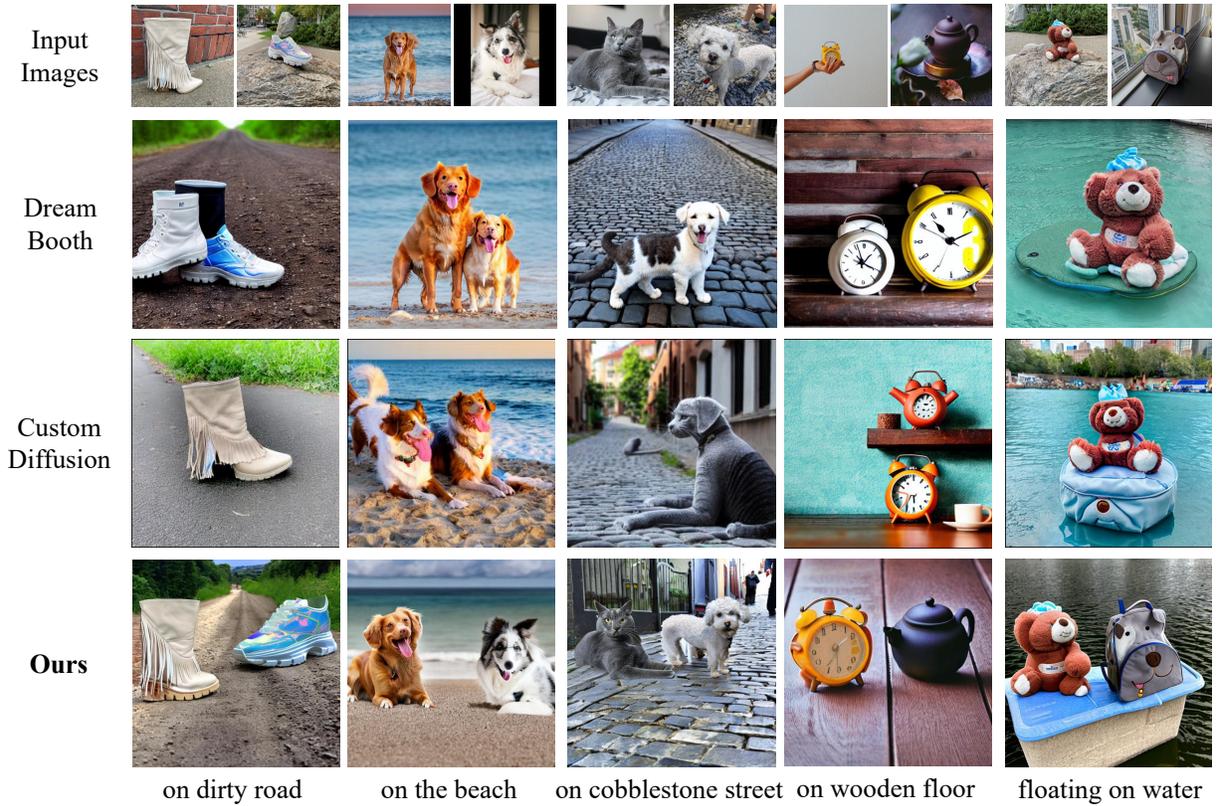}
 \caption{\small{Qualitative result for two-subject generation. All original images are from the DreamBench~\cite{ruiz2023dreambooth} dataset.}}
\label{fig:quali-multi}
\end{figure*}

\begin{figure*}[htbp]
	\centering
\includegraphics[width=0.75\linewidth]{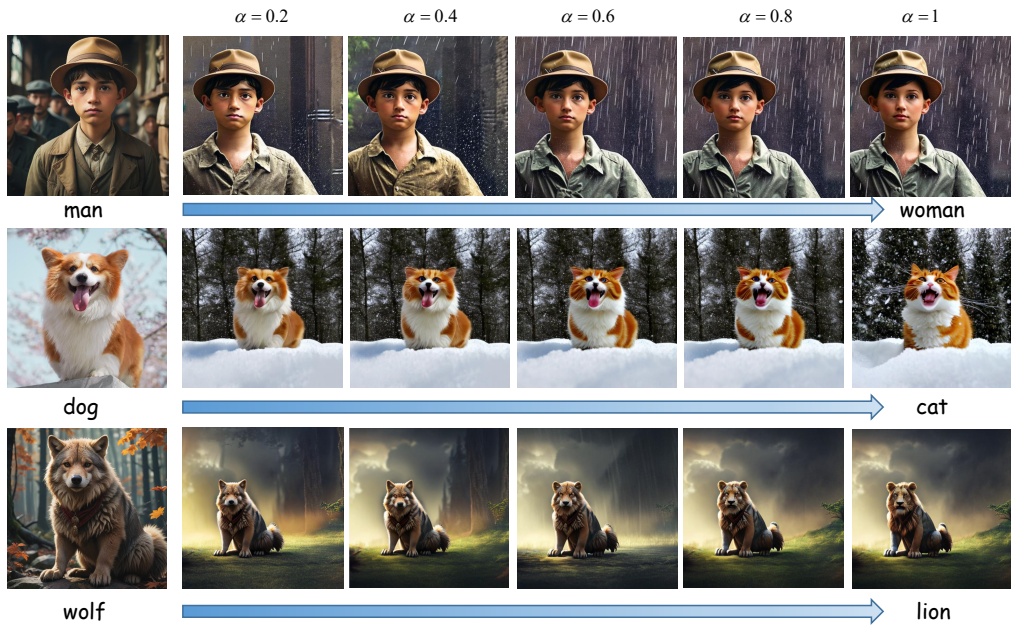}
 \caption{\small{Text-image interpolation. The prompts are followings: \textit{A man in the rain, the woman is [PH]}; \textit{A dog in the snow, the cat is [PH]}; \textit{A wolf in the forest, the lion is [PH]}.} The dog image is from the DreamBench~\cite{ruiz2023dreambooth} dataset.}
\label{fig:interpolation}
\end{figure*}

\end{document}